\pdfoutput=1

\documentclass[11pt]{article}
\usepackage{xcolor}

\usepackage[preprint]{acl}

\usepackage{times}
\usepackage{latexsym}
\usepackage{booktabs}
\usepackage{graphicx}
\usepackage{multirow}
\usepackage{booktabs}
\usepackage[T1]{fontenc}

\usepackage[utf8]{inputenc}

\usepackage{microtype}

\usepackage{inconsolata}

\usepackage{graphicx}

\usepackage{amsmath}
\usepackage{amsfonts}
\usepackage{xspace}
\usepackage{float}

\newcommand{\spiritlm}{SpiRit-LM\xspace}

\title{Interleaved Speech Language Models Latently Work In Text}

\author{Talia Sternberg \quad Gallil Maimon \quad Yossi Adi \\ The Hebrew University of Jerusalem \\ \texttt{talia.sternberg@mail.huji.ac.il}}

\begin{document}
\maketitle

\begin{figure*}[t]
    \centering
    \includegraphics[width=\textwidth]{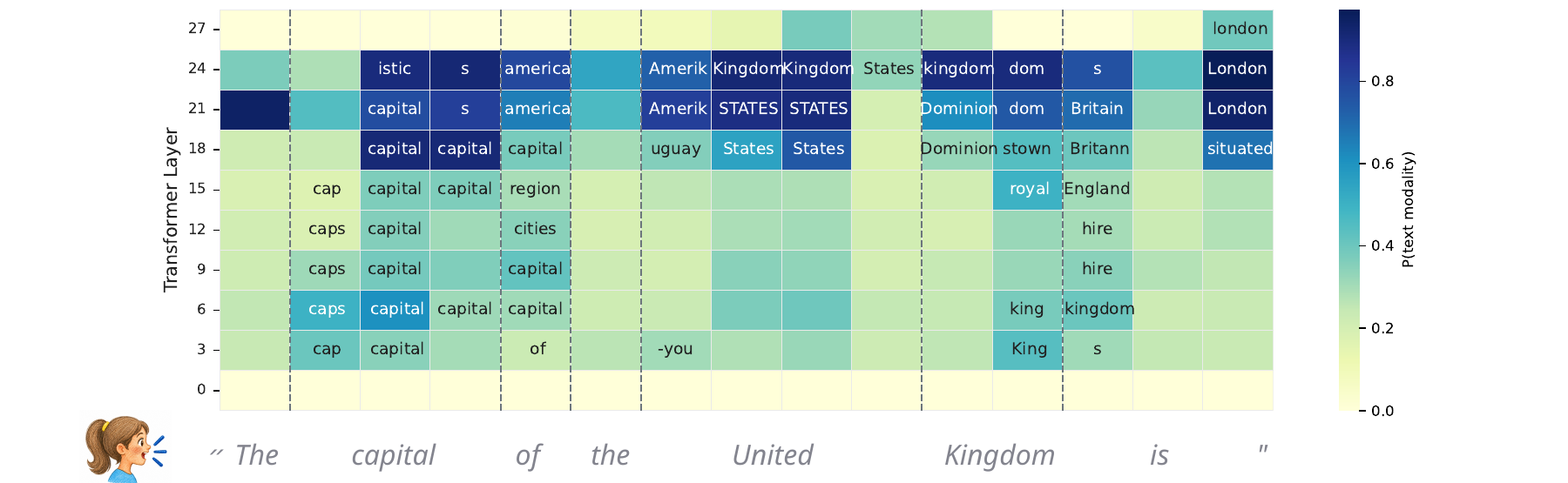}
    \caption{\textbf{Implicit transcription emerges without speech-recognition supervision.}
Logit-lens analysis of intermediate states for the spoken prompt ``The capital of the United Kingdom is...''. Cells show textual tokens probability, from light yellow for zero to dark blue for high probability. The labels show the most probable relevant textual token at each position; notably, the model predicts ``London'' although it was not spoken in the prompt.}
    \label{fig:capital_of_uk}
\end{figure*}

\begin{abstract}
Speech language models (SLMs) increasingly combine speech and text, often by
interleaving their tokens within a single sequence.
Yet how these two modalities interact in the model's latent space
remains unclear. In this work, we analyze interleaved speech--text
LMs from different model families and training configurations using
three complementary methods. 
We reveal that these models pass through an \emph{implicit latent transcription} phase in which the text token matching the spoken word becomes decodable in intermediate layers, \emph{despite not being trained for speech recognition}.
This phenomenon occurs in diverse, natural speech, and intermediate
representations also encode likely text continuations. We further
show that implicit transcription emerges most when combining text-LM pretraining and speech--text interleaving, and
that its prevalence is positively associated with spoken factual-knowledge
retrieval. Our analysis sheds light on the internal
interaction between speech and text modalities in interleaved SLMs.
\end{abstract}

\section{Introduction} \label{sec:intro}
Speech language models (SLMs) are gaining popularity as the basis for universal speech processing systems as well as dialogue models \cite{arora2026landscapespokenlanguagemodels}. Such models hold potential to reason about speech \citep{yosha2026stresstestspeechlmhandle, sakshi2025mmau}, as well as answer spoken questions or instructions \citep{nachmani2024spoken, chen2024voicebench}. However, several recent works have indicated lacking semantic and knowledge capabilities in speech-only SLMs \cite{cuervo2024scaling}.

More recently Speech LMs started integrating text data and pre-trained text LMs into speech LMs \cite{defossez2024moshi, hassid2023textually, xie2024mini}. One such method is speech-text interleaving, in which models are trained over ``code switching'' data which contains both speech and text tokens as a single stream \cite{nguyen-etal-2025-spirit, zeng2025scaling, manakul2026scaling}. \citet{maimon2025sims} demonstrated that this interleaving method improves scaling dynamics of the semantic abilities of SLMs relative to the speech-only paradigm, even when considering speech-in speech-out performance. While these methods improve speech-in speech-out capabilities, the mechanism underlying this behavior, and the internal interplay between speech and text modalities, still remain unclear.

In this work, we analyze latent dynamics between modalities using three complementary decoding methods. We use the logit lens \cite{nostalgebraist2020logitlens}, which projects intermediate hidden states into the token vocabulary, and complement it with a latent-space retrieval method \citep{krojer2026latentlensrevealinghighlyinterpretable} and supervised linear probes \citep{alain2018understandingintermediatelayersusing, belinkov2021probingclassifierspromisesshortcomings}. This analysis reveals a clear pattern in SLMs’ internal representations (Figure \ref{fig:capital_of_uk}): interleaved SLMs operate in a meaningful textual latent space within the hidden layers of the transformer. Specifically, speech-derived representations become decodable as the corresponding text transcription and as hypotheses about the next word, before being projected back into the speech-token space. This occurs despite the models not being explicitly trained for speech transcription. We refer to this phenomenon as \emph{implicit latent transcription}.

We further show in Section~\ref{sec:natural_speech} that these findings occur across diverse speakers, domains, and recording conditions - from controlled synthesized prompts to natural speech.

In Section \ref{sec:ablations}, we analyze which training decisions impact this phenomenon. Notably, we highlight that implicit latent transcription emerges most strongly when combining initialization from a trained text LM and speech-text interleaving. Conversely, models without interleaving or without text-LM initialization exhibit substantially weaker transcription signals.

Finally, we analyze how this phenomenon correlates with basic spoken factual knowledge retrieval. Our results, presented in Section~\ref{sec:correlation}, suggest that implicit transcription is positively associated with factual knowledge abilities. We also qualitatively analyze examples and find that transcriptions often build gradually over the course of a spoken word and sometimes contain acoustic errors (Section~\ref{sec:qualitative}).

Our main contributions are: (i) Showing that interleaved Speech LMs latently transcribe the current spoken word and encode hypotheses about textual continuations, while not being trained for speech recognition, with consistent evidence across three complementary 
methods. (ii) Highlighting the importance of both text LM initialization, and interleaving data in eliciting this behavior. (iii) Analyzing the relationship between implicit transcription and spoken knowledge abilities. 
The dataset, audio samples, and additional implicit-transcription
visualizations are available on our project page.\footnote{
\url{https://pages.cs.huji.ac.il/adiyoss-lab/slm_work_in_text/}}
\section{Background}
\label{sec:back}

\paragraph{Interleaved Speech Language Models} are trained on both discrete speech units and text tokens. These models commonly include three components: a speech-to-unit module, a joint speech-text LM, and a unit-to-speech module~\citep{arora2026landscapespokenlanguagemodels}. The speech-to-unit module converts raw audio into discrete units, typically using a self-supervised model such as HuBERT~\citep{hsu2021hubertselfsupervisedspeechrepresentation} and quantizing these representations. For more on speech tokenization see \citet{mousavi2025discrete}. The joint Speech LM then models sequences containing both speech units and text tokens, often initialized from a pretrained text LM~\citep{hassid2023textually}. Finally, generated speech units can be converted back into waveform audio using a unit-to-speech decoder~\citep{lakhotia-etal-2021-generative}.

We specifically focus on interleaved speech-text SLMs, where speech units and text tokens are mixed within the same sequence~\citep{nguyen-etal-2025-spirit}. Such models show promising results with elegant modeling of a single stream of tokens ~\cite{nguyen-etal-2025-spirit, maimon2025sims, zeng2025scaling}. For model training, given time-aligned transcriptions of speech samples, each word is assigned either the speech or text modality. Consecutive words with the same modality are grouped into spans: text spans are tokenized as text, while speech spans are replaced by their discrete units. This yields sequences such as \texttt{[TEXT] the monkey climbed [SPEECH] Hu14 Hu62 Hu9 \(\ldots\) Hu31 [TEXT] tree.} 

Such a modeling approach trains the SLM to generate cross-modal continuations within the same sentence, under the assumption that semantic information is transferred across modalities.

\section{Approach}
\label{sec:approach}

\subsection{Interpreting Latent Embeddings}
\label{sec:decoding_methods}
Let $h_i^{(\ell)}$ denote the hidden state at speech-token position
$i$ and layer $\ell$. We aim to understand what information is encoded in speech LM latent representations, and how this information evolves across the layers of the model. To do so,
we decode these representations using three
complementary methods.

\paragraph{Logit Lens.}
The logit lens \citep{nostalgebraist2020logitlens} applies the model's
final normalization and output projection to intermediate hidden states:
\[
p_{\mathrm{logit}}^{(\ell)}(x \mid i)
=
\operatorname{softmax}
\left(
W_{\mathrm{out}}
\operatorname{Norm}_{f}\!\left(h_i^{(\ell)}\right)
\right)_x ,
\]
where $\operatorname{Norm}_{f}$ is the model's final RMSNorm and
$W_{\mathrm{out}}$ is its learned vocabulary projection.

Although this projection is normally applied only at the final layer,
applying it throughout the network yields a layer-wise distribution over
the vocabulary and reveals which token predictions are already linearly
decodable at each depth. Since transformer predictions are formed
gradually through the shared residual stream
\citep{geva2022transformer}, this provides a direct view of how
vocabulary-aligned information emerges and changes across layers.
Logit-lens analyses have been widely used to study the evolution of
intermediate predictions
\citep{dar2023analyzing,wendler2024llamas,yang2024large,
halawi2024overthinking,neo2025towards}.

\paragraph{Latent Lens.}
The logit lens only reveals information aligned with the model's output
projection. We therefore complement it with Latent Lens
\citep{krojer2026latentlensrevealinghighlyinterpretable}, which tests whether speech hidden states
are geometrically close to text representations produced by the same
model.

We collect contextualized occurrences of the evaluation words and
random distractors from held-out Red Pajama text. At each layer, we
average the representations of all occurrences of each word to obtain
one model-specific text representation per word. We then rank the words
by their cosine similarity to each speech hidden state and treat the
top-$k$ words as the decoded candidates.

\paragraph{Linear Probes.}
We additionally use simple linear probes
\citep{alain2018understandingintermediatelayersusing, belinkov2021probingclassifierspromisesshortcomings} to test whether the
current and next words are linearly recoverable from speech hidden
states. Each probe is a single affine classifier followed by a softmax. For each model and layer, we train
separate current- and next-word probes while keeping the SLM frozen.
Each probe predicts over the $10{,}000$ most frequent
tokenizer-specific RedPajama tokens , together with all evaluation
targets, and is trained on $5M$ aligned speech-token positions
from People's Speech dataset \citep{galvez2021peoplesspeechlargescalediverse}. The probe measures whether lexical information is linearly recoverable, even when it is not aligned with the model's output head or the text representations used by the latent lens.\\\\
Across all decoding methods, we lowercase text and strip punctuation to ensure exact string matching, then evaluate on the first word-start token.

\subsection{Word-Level Analysis of Speech Latents}
\label{sec:word_level_analysis}

A spoken word spans a variable number of speech tokens, and no fixed
position within the span necessarily contains the strongest information
about its identity. Let $S(w)$ denote the speech-token positions aligned
with word $w$.

\paragraph{Modality Mass} is calculated using the logit lens, as it is the only
method that produces a probability distribution over the joint
speech-text vocabulary. Let

\[
p_{\mathrm{logit}}(i)^{(\ell)}
=
\operatorname{softmax}\!\left(
W_{\mathrm{out}}
\operatorname{Norm}_{f}\!\left(h_i^{(\ell)}\right)
\right)
\]
denote the vocabulary distribution at position $i$ and layer $\ell$.
For modality $m\in\{\mathrm{text},\mathrm{speech}\}$, we define the
word-level modality score as
\[
s_m^{(\ell)}(w)
=
\max_{i\in S(w)}
\sum_{v\in\mathcal{V}_m}
p_{\mathrm{logit}}(i)^{(\ell)}(v),
\]
where $\mathcal{V}_m$ is the corresponding modality vocabulary.

\paragraph{Word Recovery.}
For decoding method
$j\in\{\mathrm{logit},\mathrm{latent},\mathrm{probe}\}$, let
$p_j(i)^{(\ell)}$ denote its output scores at position $i$ and layer
$\ell$. These are vocabulary probabilities for the logit lens and
probe, and cosine-similarity scores over bank entries for the latent
lens.

A word is recovered at layer $\ell$ if its target label appears among
the top-$k$ candidates at any position in $S(w)$. Because different
words become decodable at different depths, we report cumulative
Recall@$k$ through layer $L$:
\begin{equation*} \begin{aligned} & \frac{1}{|\mathcal{D}|} \sum_{w\in\mathcal{D}} \mathbf{1} \Big[ \exists\,\ell\leq L,\ i\in S(w) \\ &\hspace{1.5cm} \text{s.t. }\, y_w\in\operatorname{TopK} \bigl(p_j(i)^{(\ell)},k\bigr) \Big]. \end{aligned} \end{equation*}

where $\mathcal{D}$ is the set of evaluated word instances and $y_w$
is the target label. Thus, a word is counted once it becomes decodable
at any layer up to $L$.

We use max aggregation because the signal is sparse and its position varies across words. Robustness to this choice, along with localization, layer-wise recall, and persistence, is reported in Appendices~\ref{app:span_agg} and~\ref{app:persistence}.

\paragraph{Matched-frequency baseline.}
To control for word-frequency effects, we compare each gold word with
20 distractor words sampled from the same RedPajama frequency band and
score them using the same Recall@$k$ protocol. Full sampling details are
provided in Appendix~\ref{app:matched_frequency_baseline}.

\subsection{Evaluation Data}

We test whether speech--text interleaving enables models to access factual knowledge from speech. Because the factual content of large speech corpora is unknown, we compare text-pretrained interleaved models against speech-only and randomly initialized baselines. Better spoken fact completion by the interleaved models would suggest that joint training supports cross-modal acquisition or retrieval of factual knowledge.

Existing speech benchmarks do not offer controlled, single-word fact-completion tasks, while standard question-answering datasets often allow multiple valid answers. We therefore construct a diagnostic dataset of 282 short prompts with a single unambiguous completion, such as \textit{``The capital of France is ...''}. The examples span colors, days and months, object functions, capital cities, arithmetic, and other common-sense categories. We synthesize the prompts with Kokoro-82M \citep{hexgrad_2025} and obtain word-level alignments using Whisper large-v3 \citep{radford2023whisper}.This controlled format enables precise analysis of factual retrieval. Statistics and examples appear in Table~\ref{tab:dataset_statistics}, and the dataset is \href{https://huggingface.co/datasets/slprl/common-sense-facts-audio}{publicly available}. We additionally validate our main findings on natural speech corpora spanning diverse speakers, domains, and recording conditions, see \ref{sec:natural_speech}.

\paragraph{Knowledge Evaluation in Speech LMs.}
We then use the above dataset for factual knowledge evaluation. Each true fact is paired with a matched counterfactual example by replacing the correct answer with an incorrect one, chosen randomly from the same category, i.e \textit{'The color of a banana is yellow'} Vs. \textit{'The color of a banana is red'}. We score a model as correct when it assigns higher likelihood to the true fact than to the wrong fact. 
\[
\log p(\text{fact}) > \log p(\text{counterfactual fact}).
\]
We report the percentage of examples for which this condition holds.
Such likelihood based evaluation is commonly used for Speech LMs \citep{hassid2023textually, maimon2025salmonsuiteacousticlanguage}.

\subsection{Experimental Setup}
\label{sec:setup}

We analyze interleaved Speech LMs based on Llama~3.2-3B
\citep{dubey2024llama} and Qwen~2.5
\cite{qwen2025qwen25technicalreport}, trained on mixtures of speech,
text, and speech--text interleaved data \cite{maimon2025sims}. Model and
training details are provided in Appendix~\ref{app:exp}.

To isolate the factors underlying implicit transcription, we train
controlled Llama~3.2-3B models while varying initialization and data
composition. Models are initialized either from a pretrained text LM
(\textsc{P}) or randomly (\textsc{R}), and trained on speech only
(\textsc{S}), balanced speech and text (\textsc{ST}), or speech, text,
and interleaved data. In the latter setting, interleaved examples
constitute $1/3$, $2/3$, or $5/6$ of the training tokens, denoted
\textsc{I-1/3}, \textsc{I-2/3}, and \textsc{I-5/6}, respectively. Importantly, all models are trained only with next-token prediction and receive no ASR or transcription loss. Time-aligned transcripts are used only to construct the interleaved sequences; the model is never trained to map a speech span directly to its corresponding transcript.
\section{Results}
\label{sec:res}

\subsection{Speech LMs Operate In Text Latent Domain}
\label{sec:work_in_text}
\begin{figure}[t]
    \centering
    \includegraphics[width=1\linewidth]{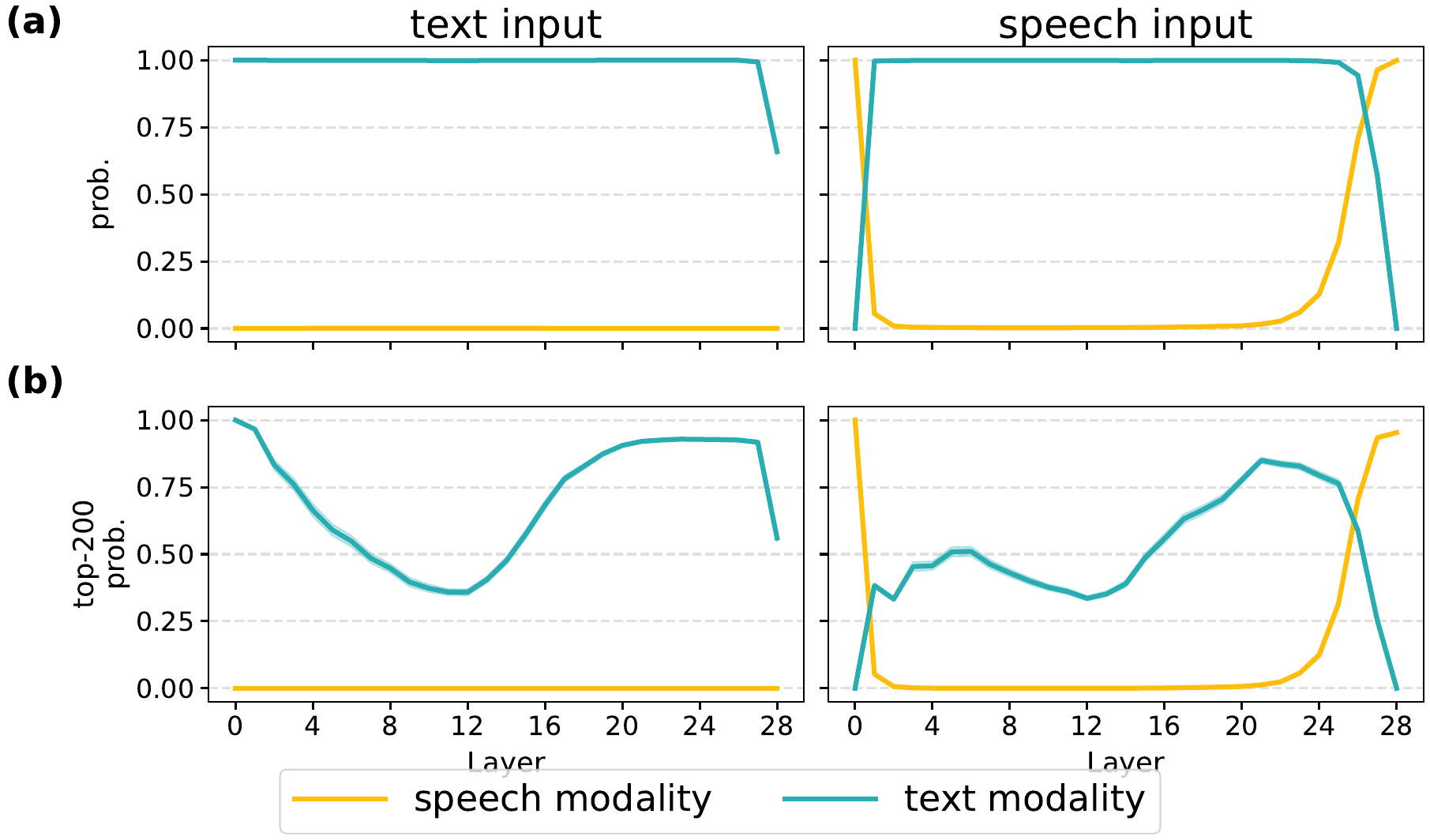}  
    \caption{\textbf{Speech LMs operate in text}.
    Modality distribution of inner state Logit Lens. (a) Sum of probabilities over all speech tokens and text tokens respectively. (b) Same but only considering top 200 tokens.}
    \label{fig:probabilities}
\end{figure}

\begin{figure*}[t]
    \centering
    \includegraphics[width=\textwidth]{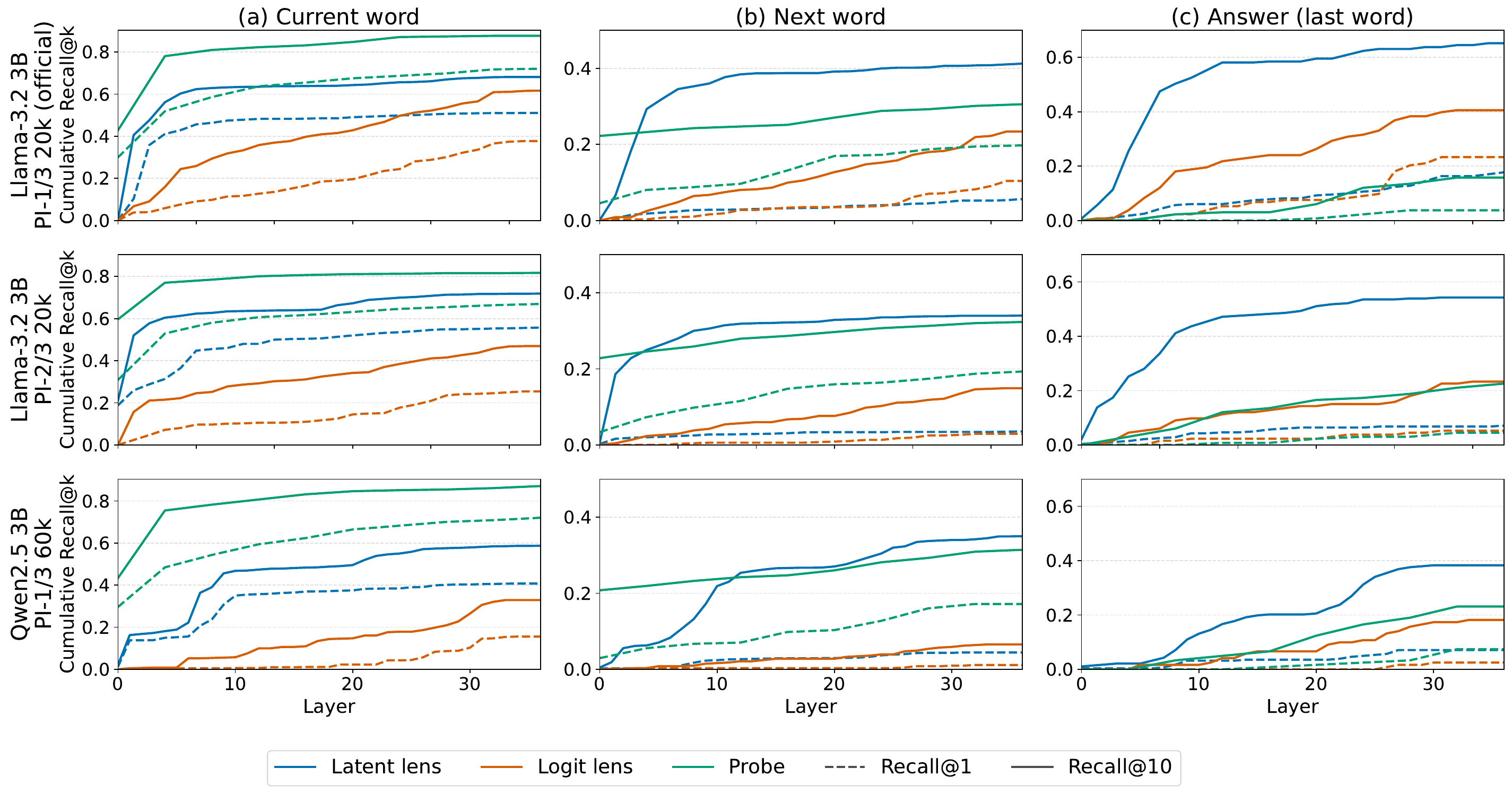}
    \caption{
\textbf{Implicit transcription and textual continuation are recoverable with complementary decoding methods. }
Cumulative $Recall@1$ (dashed) and $Recall@10$ (solid) across layers for the current word, next word, and final answer, using the logit lens, latent lens, and linear probes. Current-word information is recovered most strongly, while next-word and answer information are weaker but remain decodable across models and methods. Full logit lens Recall@k curves and matched-frequency baselines are shown in Figure \ref{fig:transcription_logit_lens}; corresponding results for the other methods are reported in Appendix~\ref{app:cum_res}. The baselines are substantially lower and are omitted here for clarity.
} 
\label{fig:transcription_all}
\end{figure*}
We first examine whether speech representations remain in the
speech-token domain throughout the network. Using the modality-mass
measure defined in Section~\ref{sec:word_level_analysis}, we compare
the logit-lens probability assigned to speech and text tokens across
layers.

Figure~\ref{fig:probabilities}a shows a clear three-stage pattern for
SIMS-Llama3.2-3B. Early layers are dominated by speech tokens
(layers $0$--$2$), intermediate layers shift strongly toward text
tokens ($2$--$25$), and the final layers return to the speech-token
domain ($26$--$28$). Thus, speech inputs pass through a text-like
intermediate regime before being mapped back to speech for generation.

For text inputs, the distribution remains concentrated on text tokens
throughout the network, indicating that this transition is specific to
speech. Because the text vocabulary is substantially larger, we repeat
the analysis using only the top-$200$ tokens
(Figure~\ref{fig:probabilities}b). The same pattern persists, showing
that the intermediate shift toward text is not explained solely by
vocabulary size.

\subsection{Speech LMs Implicitly Transcribe Words}
\label{sec:implicit_transcription}

A shift toward the text vocabulary does not by itself establish that
speech representations encode the spoken content in text. We therefore
test whether the transcription of each spoken word can be recovered
from its aligned speech-token hidden states using the logit lens,
latent lens, and linear probes.

Figure~\ref{fig:transcription_all}a shows that current-word transcription is
strongly recoverable across all three models and decoding methods.
Recovery is substantial even at Recall@1 and increases further at
Recall@10. For Llama-3.2 3B PI-1/3, cumulative
Recall@10 reaches approximately 62\% with the logit lens, 69\% with
the latent lens, and over 80\% with the linear probe. The effect also
generalizes to Llama-3.2 3B PI-2/3 and Qwen2.5 3B PI-1/3,
although with lower overall recall. Thus, the result is neither
model-specific nor an artifact of a particular decoder: speech hidden
states align with the correct vocabulary token, lie close to
contextualized text representations of the same word, and make its
identity linearly recoverable.

Recovery accumulates primarily through the intermediate layers,
consistent with the text-modality shift observed in
Section~\ref{sec:work_in_text}. Non-cumulative layer-wise analyses further
show that this pattern is not an artifact of cumulative aggregation:
across decoding methods, transcription recall is concentrated in the
intermediate layers and decreases toward the output
(Appendix~\ref{app:persistence},
Figures~\ref{fig:transcription_recall_all_by_layer} - \ref{fig:transcription_recall_all_by_layer_probe}).

\begin{figure*}[t]
    \centering
    \includegraphics[width=\textwidth]{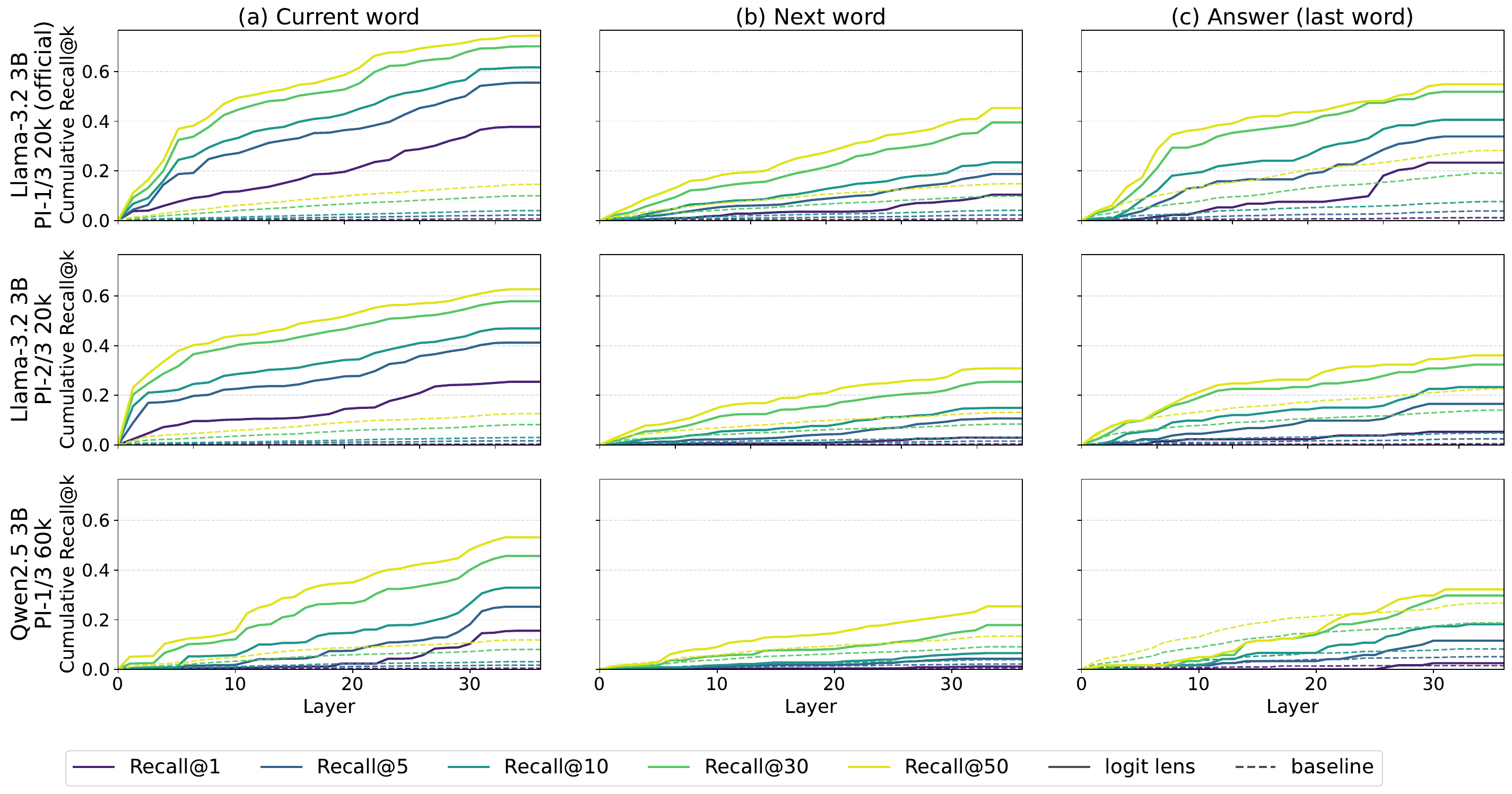}
    \caption{
\textbf{Full cumulative recovery using the logit lens.}
Cumulative Recall@\(k\) across layers for the current word, next word,
and final answer. Solid curves show recovery of the gold target, while
dashed curves show the matched-frequency control. Current-word
transcription is strongly recoverable across interleaved models, while
next-word and answer information are weaker but remain above the
frequency-matched baseline.
}
\label{fig:transcription_logit_lens}
\end{figure*}

Additional analyses further characterize this transcription signal.
Using the logit lens, we find that recovery remains substantial under
stricter within-span selectors and is typically localized to only a few
speech-token positions
(Appendix~\ref{app:span_agg}). Moreover, among words recovered
at least once, a substantial subset remains recoverable across multiple
consecutive layers
(Appendix~\ref{app:persistence}).

Matched-frequency controls are substantially lower for all three decoding methods. Figure~\ref{fig:transcription_logit_lens} shows the full logit lens Recall@k curves, demonstrating that the same pattern holds across different values of $k$. Corresponding full Recall@k curves for the latent lens and linear probes are reported in Appendix ~\ref{app:cum_res}.
Because the models are trained only with next-token prediction and receive no ASR or transcription objective, these results show that their intermediate speech representations implicitly encode the corresponding textual transcription. We refer to this as \emph{implicit latent transcription}.

\begin{table*}[t]
\centering
\small
\setlength{\tabcolsep}{5pt}
\begin{tabular}{ll|cc|cc|cc}
\toprule
\textbf{Model}
& \textbf{Data}
& \multicolumn{2}{c|}{\textbf{Logit lens}}
& \multicolumn{2}{c|}{\textbf{Latent lens}}
& \multicolumn{2}{c}{\textbf{Linear probe}} \\
& & \textbf{Cur} & \textbf{Next}
& \textbf{Cur} & \textbf{Next}
& \textbf{Cur} & \textbf{Next} \\
\midrule

\multirow{2}{*}{Llama-3.2 3B PI-1/3 20k (official)}
& Natural    & 71.7 & 27.1 & 40.3 & 20.0 & 89.4 & 57.4 \\
& Controlled & 61.9 & 23.9 & 68.1 & 41.2 & 87.7 & 30.6 \\
\midrule

\multirow{2}{*}{Llama-3.2 3B PI-2/3 20k}
& Natural    & 57.7 & 17.4 & 51.7 & 28.2 & 87.7 & 58.5 \\
& Controlled & 49.0 & 15.7 & 71.9 & 34.0 & 81.7 & 32.3 \\
\midrule

\multirow{2}{*}{Qwen2.5 3B PI-1/3 60k}
& Natural    & 40.8 & 16.5 & 43.6 & 28.9 & 88.0 & 56.7 \\
& Controlled & 32.9 &  6.5 & 58.7 & 35.0 & 87.1 & 31.4 \\
\bottomrule
\end{tabular}

\caption{
\textbf{Implicit transcription and textual continuation generalize to
natural speech.}
Cumulative Recall@10 on natural speech and the controlled common-sense
dataset using the logit lens, latent lens, and linear probes.
Natural-speech results are aggregated across seven corpora.
Cur and Next denote recovery of the current and next words,
respectively. Corpus-level results and matched-frequency controls are
reported in Appendix~\ref{app:natural_speech}.
}
\label{tab:natural_summary}
\end{table*}

\begin{table*}[t]
\centering
\small
\begin{tabular}{l|cccc|ccc}
\toprule
\textbf{Model} 
& \textbf{Text pre-trained}
& \textbf{Text}
& \textbf{Inter.}
& \textbf{Inter. frac.}
& \textbf{Cur}
& \textbf{Next}
& \textbf{Ans} \\
\midrule
SIMS Llama-3.2 PI-1/3 (official) & \checkmark & \checkmark & \checkmark & \(1/3\) & \textbf{\underline{61.88}} & \textbf{\underline{23.91}} & \textbf{\underline{41.60}} \\
\midrule
Llama-3.2 PI-1/3 (ours)              & \checkmark & \checkmark & \checkmark & \(1/3\) & \textbf{48.75} & \textbf{15.31} & \textbf{26.40} \\
Llama-3.2 PI-2/3
       & \checkmark & \checkmark & \checkmark & \(2/3\) & \textbf{49.06} & \textbf{15.78} & 24.00 \\
Llama-3.2 PI-5/6
        & \checkmark & \checkmark & \checkmark & \(5/6\) &  2.66 &  1.72 &  4.00 \\
Llama-3.2 PST
        & \checkmark & \checkmark & x          & --   &  3.75 &  3.28 &  0.00 \\
Llama-3.2 PS          & \checkmark & x          & x          & --   &  0.16 &  0.00 &  0.00 \\
Llama-3.2 RST
            & x          & \checkmark & x          & --   &  5.78 &  6.72 &  4.80 \\
Llama-3.2 RI-1/3
      & x          & \checkmark & \checkmark & \(1/3\) &  2.19 &  5.78 &  7.20 \\
\bottomrule
\end{tabular}
\caption{
Recall@10 for different Speech LMs, i.e the percentage of examples in which the correct current word (Cur), next word (Next), or answer word (Ans) appears among the top-10 predicted tokens across the relevant spoken word in any transformer layer.
}
\label{tab:cur_next_ans_results_third}
\end{table*}

\subsection{Models Predict the Continuation in Text}
\label{sec:think} 
We next ask whether the intermediate text-like representations encode
only the current spoken word or also its expected textual continuation.
Using the same protocol, we replace the current-word target with either
the next word in the sentence or, at the final prompt position, the
correct factual answer.

Figure~\ref{fig:transcription_all}b shows that next-word information is recoverable
across models and decoding methods, although less strongly than the
current transcription. This weaker recovery is partly expected because
the next word is not always uniquely determined. For example, after
``United,'' both ``Kingdom'' and ``States'' may be plausible
continuations. Lower recall may therefore reflect target ambiguity as
well as a weaker predictive signal.

We consequently evaluate the final prompt word against the factual
answer, which provides a less ambiguous continuation target. As shown
in Figure~\ref{fig:transcription_all}c, answer information is recoverable across
all three methods, particularly for Llama-3.2 3B PI-1/3 20k
(official). Figure~\ref{fig:capital_of_uk} illustrates one such example,
where ``London'' becomes decodable although it was not spoken in the
prompt. Additional examples are provided in
Appendix~\ref{app:trans_ex}.

Matched-frequency controls are substantially lower for both continuation targets, with the same pattern holding across $k$ values for the logit lens (Figure~\ref{fig:transcription_logit_lens}); corresponding results for the other methods are reported in Appendix~\ref{app:cum_res}. 

Together, these results
show that intermediate speech representations encode not only the
current transcription, but also likely textual continuations. However,
we do not observe a consistent transcription-then-prediction ordering
across the dataset, suggesting that the two signals may overlap rather
than form a fixed sequential process.

\subsection{Validation on Natural Speech}
\label{sec:natural_speech}

We test whether implicit transcription generalizes to natural speech using $2,100$ utterances sampled evenly from seven corpora spanning diverse speaking styles and recording conditions.

Table~\ref{tab:natural_summary} reports cumulative Recall@$k$ alongside results on the controlled common-sense dataset; corpus-level results are provided in Appendix~\ref{app:natural_speech}.
The same qualitative patterns are observed across settings and corpora. Thus, both effects extend beyond synthesized prompts.

\subsection{What Impacts Implicit Transcription Prevalence?}
\label{sec:ablations}

\begin{figure*}[t]
    \centering
    \includegraphics[width=\linewidth]{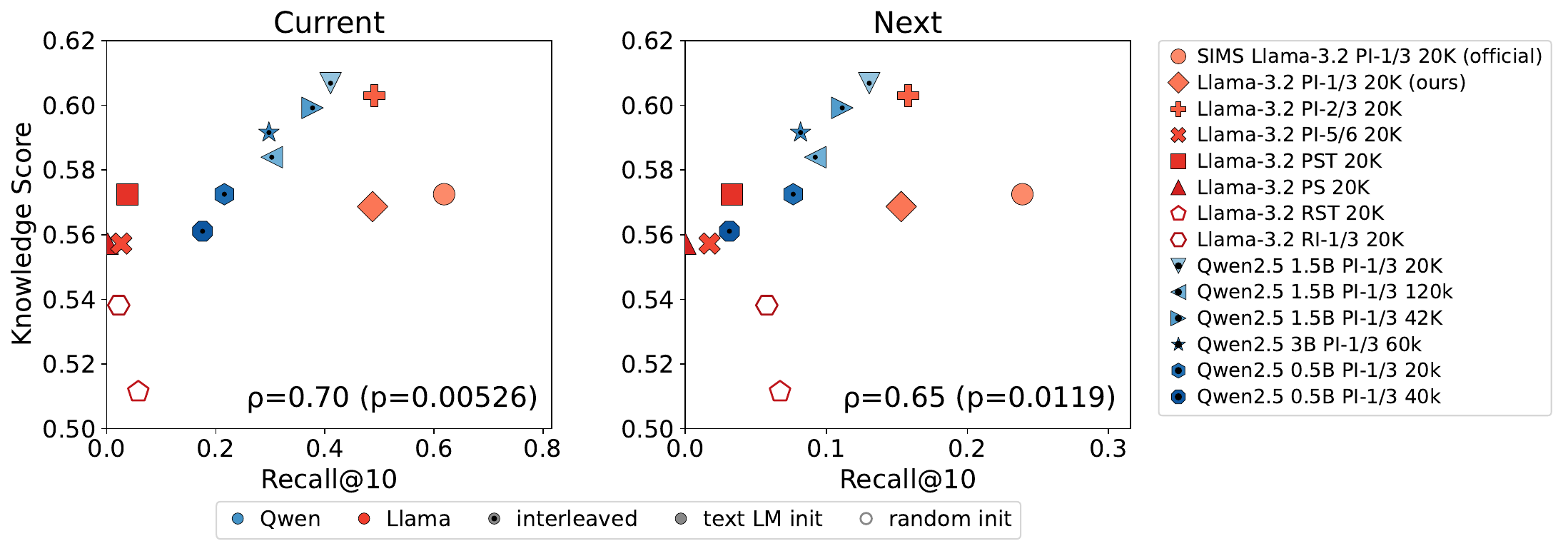}  
    \caption{
    \textbf{Implicit transcription correlates with factual knowledge retrieval.}
    Across model variants, knowledge accuracy increases with current-word (left) and next-word (right) transcription Recall@10, aggregated across aligned speech-token positions and layers.
    }
    \label{fig:corr}
\end{figure*}

We next ask whether implicit transcription depends on text-LM initialization, speech--text interleaving, or their combination. We compare Llama-3.2-3B variants spanning pretrained and random initialization, with and without interleaving, and different interleaving fractions.

Because the three decoding methods show consistent qualitative patterns
on the representative models, we use the more computationally efficient
logit lens for the full model sweep in this section and
Section~\ref{sec:correlation}. We report final-layer cumulative
Recall@10: the percentage of targets recovered at any position in the
aligned word span and at any layer up to the model output.

As shown in Table~\ref{tab:cur_next_ans_results_third}, for the setting of $k=10$, implicit transcription is strongest in text-pretrained models trained with interleaved speech-text data. Both the official and our Llama-3.2-PI-1/3 variants show clear transcription signals when interleaved data fraction was at at most $2/3$. In contrast, models trained without interleaved data show substantially weaker transcription, even when text is included in the training mixture. Similarly, randomly initialized models show only weak transcription signals, including when trained with interleaved data. These results suggest that, in this setting, neither exposure to text tokens nor interleaved data alone is sufficient; implicit transcription emerges most clearly from their combination with a pretrained textual prior. Results for different values of $k$ show the same pattern  (Tables \ref{tab:cur_next_ans_results_k1}, \ref{tab:cur_next_ans_results_k5}, \ref{tab:cur_next_ans_results_k30} in Appendix \ref{app:trans_recall}).

The fraction of interleaved data also affects the strength of implicit transcription. Models trained with lower or intermediate interleaved-token fractions, such as $1/3$ and $2/3$, show strong implicit transcription, whereas the model trained with a much higher interleaved-token fraction, $5/6$, shows a substantially weaker signal. This suggests that the effect is not simply monotonic in the amount of interleaved data: excessive interleaving may shift the training distribution or reduce exposure to pure modality-specific contexts.

Overall, these ablations suggest that implicit transcription is not merely a byproduct of using a unified vocabulary or exposing the model to text tokens. Rather, it appears most clearly when a text-pretrained model receives sufficient interleaved speech-text supervision to align spoken units with textual representations.

\subsection{Does Implicit Transcription Correlate With Factual Knowledge?}
\label{sec:correlation}

We test whether stronger implicit transcription is associated with
better factual-knowledge retrieval across the official SIMS models and
our variants from Section~\ref{sec:ablations}. These models span
Llama 3.2 and Qwen 2.5 architectures, model sizes, initializations,
training budgets, and data mixtures. Following
Section~\ref{sec:ablations}, we use the same final-layer cumulative
logit-lens Recall@10 for the cross-model correlation analysis. For each
model, we compare current- and next-word recovery with accuracy on our
common-sense knowledge benchmark.

We test the monotonic association between transcription recall and factual accuracy using two-sided Spearman rank correlations. Both  measures are positively correlated with factual accuracy: current-word recall yields $\rho=0.70$ ($p=0.0053$), while next-word recall yields $\rho=0.65$ ($p=0.0119$). Both correlations remain significant under the Bonferroni-adjusted threshold of $\alpha=0.025$.

However, this across-model correlation does not establish sufficiency or causality. Randomly initialized models, for example, show non-zero transcription scores despite low knowledge accuracy, and aggregation may obscure differences across layers and positions. We therefore interpret this as an association rather than a mechanistic explanation.

\subsection{Qualitative Analysis}
\label{sec:qualitative}
As part of a qualitative analysis of implicit transcription, we observed recurring phenomena that may provide broader context for this behavior. We present these findings as exploratory observations, and leave a broader and more systematic investigation of their scope and implications to future work.
For this analysis, we also consider single word utterances, synthesized using the same approach as our evaluation data, to provide a simpler setting for interpretation.

\paragraph{Implicit Transcription Builds Gradually.}
Implicit transcription sometimes appears to emerge incrementally over the course of a spoken word. For example, the logit lens first decodes \emph{white} as "why" (Figure \ref{fig:transcription_white}), \emph{lime} as "lie" (Figure \ref{fig:transcription_lime}), \emph{kingdom} as "king" (Figure \ref{fig:transcription_uk_full}), and \emph{Pakistan} (Figure \ref{fig:transcription_pakistan}) as "pack", before later converging to the full word. These examples suggest that the model may construct textual representations from partial acoustic evidence, updating them as more of the word becomes available. More examples can be found in Appendix \ref{app:trans_ex}. As a result, intermediate representations can temporarily correspond to acoustically plausible but semantically unrelated words. Such transient mismatches may contribute to the broader speech - text representation gap, and we leave a systematic investigation of this effect to future work.

\paragraph{Implicit Transcription Has Errors.}
We also observe incorrect but acoustically similar implicit transcriptions. For example, \emph{lime} is decoded as ``line'' (Figure \ref{fig:transcription_lime}, Appendix \ref{app:trans_ex}).
Unlike the gradual prefix effect above, these cases suggest that the model occasionally settles on a text interpretation that is acoustically close to the input but is incorrect. Such errors highlight that the implicit transcription process may introduce ambiguity or noise into the speech-derived text representation.
\section{Related Work}\label{sec:related}

\paragraph{Interpreting Latent Computation in LMs.}
Several works suggest that multilingual LLMs share information across languages through English-centric latent representations; \citet{wendler2024llamas} show this in Llama-2, and related English-mediated processing or steering effects are reported by \citet{zhao2024largelanguagemodelshandle,mahmoud2025improvingmultilinguallanguagemodels}. Related multi-modal work proposes a similar semantic-hub view, where representations from different modalities align in intermediate layers~\cite{wu2025semantichubhypothesislanguage}. Closest to our analysis, \citet{neo2025towards} apply logit lens to visual-token hidden states in LLaVA, showing that image representations can become decodable in the language vocabulary despite not corresponding to discrete text tokens. Complementarily, \citet{nikankin2026same} further show that vision and text may rely on distinct circuits, with alignment emerging only in late layers. Together, these works suggest that cross-modal reasoning requires non-textual inputs to enter a shared workspace early enough. However, they focus on multilingual text or vision-language models, whereas we study whether such a text workspace emerges during speech processing.

\paragraph{Analyzing Speech-Text LMs.}
Several studies have tried to interpret speech-text LMs by analyzing the sources of the modality gap: the mismatch between speech and equivalent text inputs. Recent work suggests that this gap is partly caused by the different structure of the two modalities; for example, redundant speech tokenization can lead to diffuse attention and weaker decision sharpening compared to text~\cite{hsu2026anatomymodalitygapdissecting}. In \spiritlm ~\cite{nguyen-etal-2025-spirit}, layer-wise analyses are used to explain why interleaving speech and text is beneficial: the two modalities become increasingly geometrically aligned across layers, suggesting that interleaving helps induce a shared latent structure that may reduce the speech-text gap. Related studies further measure speech-text alignment across layers~\cite{mousavi2025alasmeasuringlatentspeechtext,xiang2025understandingmodalitygapempirical}. Other work finds that audio LMs often rely on transcript-like information, behaving similarly to ASR$\rightarrow$LLM cascades on text-sufficient tasks~\cite{chen2025audiollmsreallylisten,billa2026cascadeequivalencehypothesisspeech}.
Recent mechanistic studies of ASR models further analyze how transcript tokens emerge across layers, using decoder logit-lens analyses, probing, and activation patching to trace acoustic and semantic information during transcription~\cite{lioubashevski2024lookingtop1transformersdetermine,glazer2025transcriptionmechanisticinterpretabilityasr}.
Nevertheless, these works focus on audio-to-text models or static speech-text alignment. In contrast, we study a generative speech-text LM that both consumes and produces speech tokens, and ask whether speech processing internally passes through a textual workspace despite the model not being explicitly trained for speech-to-text transcription.

\section{Discussion \& Conclusion} \label{sec:con}

Our work, clearly indicates that Speech LMs implicitly move to a text latent space in which the corresponding text transcription and hypotheses about the next word become decodable before being projected back to speech tokens. We observe this behavior consistently across three complementary decoding methods and show that it generalizes from controlled synthesized prompts to natural speech. We also demonstrate that this positively correlates with spoken fact knowledge abilities.

This work immediately opens up future research question, most excitingly can this behavior be optimized directly leading to overall better performance. However, acting on these insights explicitly could also have negative sides. Specifically, we left for future work the analysis of the effect of implicit transcription on acoustic abilities. As acoustic abilities are a key motivation for modeling speech directly and avoiding \emph{explicit} transcription, this poses a key question.

Another interesting question that arises from this study is ``what limits the spoken abilities relative to text and leads to a modality gap?''. Given that SLMs latently work in text, one could expect a small modality gap and yet this is not the case. Future work could further analyze if this has to do with transcription error, compute ``wasted'' on transcription, or lack of temporal compression making language reasoning more challenging.

\subsection*{Limitations}
Our work shows that implicit transcription emerges in intermediate layers, but does not identify the precise mechanism that computes it, such as the specific heads, layers, or pathways involved. More generally, our decoding analyses establish that information is accessible in intermediate representations, but not that it is causally used by the model. The logit lens is limited to information aligned with the model's output projection, while linear probes demonstrate linear decodability through a learned classifier but do not establish that the model itself uses this information during generation. Although agreement across complementary methods strengthens the evidence for a text-like intermediate representation, it does not establish its causal role. Moreover, although transcription-like signals are positively associated with factual knowledge retrieval from speech, the correlation reaches only up to $\rho=0.70$ and is not sufficient to explain all variation across models. We do not causally test this relationship by intervening on models to increase implicit transcription and measuring whether knowledge retrieval improves, and leave this analysis for future work.

\paragraph{Acknowledgments.} This research work was supported by Israel Science Foundation (ISF), grant number $2049/22$. 

\bibliography{refs}

\newpage
\clearpage
\appendix
\section{Appendix}
\label{sec:appendix}

\subsection{Common Sense Dataset}
\label{app:dataset}
Table~\ref{tab:dataset_statistics} reports statistics for the different subsets of our evaluation dataset, along with one representative example from each subset.

\subsection{Matched-Frequency Baseline}
\label{app:matched_frequency_baseline}

For each gold word, we sample 20 control words from the same frequency
band in a fixed RedPajama word-frequency table based on the Llama-3.2
tokenizer. The same reference is used for all models. Controls are
restricted to words in the latent-lens bank and exclude the gold word
and all evaluation words. If a band contains too few candidates, we
expand the pool to adjacent bands and then to all alphabetic bank words.

Controls are scored using the same span-level and cumulative
Recall@$k$ procedure as the gold words. We report the mean recall across
the 20 samples with 95\% confidence intervals.

\subsection{Experimental Setup}
\label{app:exp}

All Llama-3.2 variants are trained with the same architecture and optimization setup, enabling controlled comparison across initialization and data composition. Matching the original SIMS paper, all models are trained with a next-token prediction objective, sequence length 2048, an effective batch size of 32 samples, and 20K training steps. Training is performed with \textsc{SLAMKit}~\citep{maimon2025slamming}.

\paragraph{Interleaving.}
We follow \citet{zeng2025scaling}, sampling speech-segment lengths from
a Poisson distribution with $\lambda=10$ until speech spans cover
$\eta=0.3$ of the words. Time-aligned transcripts are used only to
identify word boundaries and assign each span to either speech or text.
Thus, each word appears in only one modality within an interleaved
sequence, rather than as a parallel speech--transcript pair. The models
are trained solely with next-token prediction, without an ASR loss or
direct speech-to-text alignment objective.

\paragraph{Training Data}
For speech training data, we follow \citet{maimon2025sims} and use the same English speech mixture: LibriSpeech \citep{panayotov2015librispeech}, LibriLight \citep{Kahn_2020}, VoxPopuli \citep{wang2021voxpopulilargescalemultilingualspeech}, TED-LIUM \citep{Hernandez_2018}, People's Speech \citep{galvez2021peoplesspeechlargescalediverse}, SWC \citep{KHN16.518}, and synthetic sTinyStories \citep{maimon2025slamming}. Text-only data is taken from RedPajama \citep{weber2024redpajama}, filtered with Gopher rules \citep{rae2021scaling}. Speech is represented with discrete HuBERT-style units \citep{hsu2021hubertselfsupervisedspeechrepresentation}: mHuBERT features are quantized using a 500-unit k-means codebook. Text is tokenized with the Llama~3.2 tokenizer. In interleaved configurations, speech and text spans are mixed using Whisper large-v3-turbo alignments \citep{radford2023whisper}.

\subsection{Additional Results}

\subsubsection{Span Aggregation and Localization Robustness}
\label{app:span_agg}

A spoken word corresponds to a variable-length span of speech tokens, and the position at which its transcription becomes decodable may differ across words. Our main analysis therefore uses any-position aggregation: a word is considered recovered if its target text token appears among the top-$k$ predictions at any position within its aligned speech span. Here, we test whether our findings depend on this choice and quantify how localized the transcription signal is within each span.

Let $S(w)=(i_1,\ldots,i_m)$ denote the ordered speech-token positions aligned with word $w$, and let $p_i^{(\ell)}$ denote the logit-lens distribution at position $i$ and layer $\ell$. We evaluate the following span selectors:

\begin{itemize}
\item \textbf{Max over position:} the word is recovered if its target token appears among the top-$k$ predictions at any position in $S(w)$.

\item \textbf{Final 25\%:} recovery is evaluated over only the final $\lceil m/4\rceil$ positions in the span.

\item \textbf{Final token:} recovery is evaluated only at the final position $i_m$.

\item \textbf{First token:} recovery is evaluated only at the first position $i_1$.

\item \textbf{Mean positional recall:} we compute the fraction of positions in the span at which the target token appears among the top-$k$ predictions. For example, if the target appears at two of five positions, the score is $2/5$, whereas any-position aggregation assigns a score of one.
\end{itemize}

These variants differ only in how recovery is aggregated across positions within a word span. 

\begin{table}[t]
\centering
\small
\resizebox{\columnwidth}{!}{
\begin{tabular}{lccc}
\toprule
\textbf{Span selector}
& \textbf{Llama PI-1/3}
& \textbf{Llama PI-2/3}
& \textbf{Qwen PI-1/3} \\
\midrule
Max over positions & 61.7 & 46.9 & 32.9 \\
Final 25\% & 38.6 & 27.8 & 14.7 \\
Final token & 33.9 & 24.6 & 9.2 \\
First token & 9.2 & 5.8 & 1.8 \\
Mean over positions & 22.9 & 16.6 & 12.1 \\
\bottomrule
\end{tabular}
}
\caption{Cumulative Recall@10 under alternative within-span selectors.
Although any-position aggregation is the
most sensitive, implicit transcription remains detectable when the
analysis is restricted to the final quarter or final token of each
word span.}
\label{tab:span_selectors}
\end{table}

As shown in Table~\ref{tab:span_selectors}, max over position aggregation yields the highest recall, but the effect remains substantial under stricter selectors. For the official Llama PI-1/3 model, Recall@10 decreases from $61.7$\% using any position to $38.6$\% using only the final quarter of the span and $33.9$\% using only the final token. The low recall at the first token is consistent with the model requiring additional acoustic context before the word becomes decodable, although imperfect boundaries and brief silences may also contribute. Mean positional Recall@10 for the official Llama PI-1/3 model is $22.9$\%. The gap between the two metrics indicates that transcription is often decodable somewhere within a word span, but is not uniformly present across all of its positions.

To quantify this localization directly, we define the number of successful positions for word $w$ at layer $\ell$ as

\begin{equation*}
n_{\mathrm{hit}}(w,\ell,k)
=
\sum_{i \in S(w)}
\mathbf{1}
\left[
y_w \in \operatorname{TopK}\left(p_i^{(\ell)},k\right)
\right]
\end{equation*}

where $S(w)$ is the set of speech-token positions aligned with word $w$, $p_i^{(\ell)}$ is the logit-lens distribution at position $i$ and layer $\ell$, and $y_w$ is the target text token.

We evaluate this statistic over word--layer pairs containing at least one top-$k$ hit. At $k=10$, the target token appears in at most two positions in $82.2$\% of successful word--layer pairs, while the median aligned word span contains $6$ speech tokens. Thus, the transcription signal is typically concentrated in a small subset of the aligned span rather than distributed uniformly across it.

Overall, max-over-position aggregation increases sensitivity to localized signals whose position varies across words, but does not create the implicit-transcription effect. The signal remains clearly detectable under stricter fixed-position selectors, while the mean positional and localization analyses confirm that it is generally concentrated at only a few positions within each speech span.

\subsubsection{Cumulative Recall Results Over Models and Decoding Methods}
\label{app:cum_res}

\begin{figure*}[t]
    \centering
    \includegraphics[width=\textwidth]{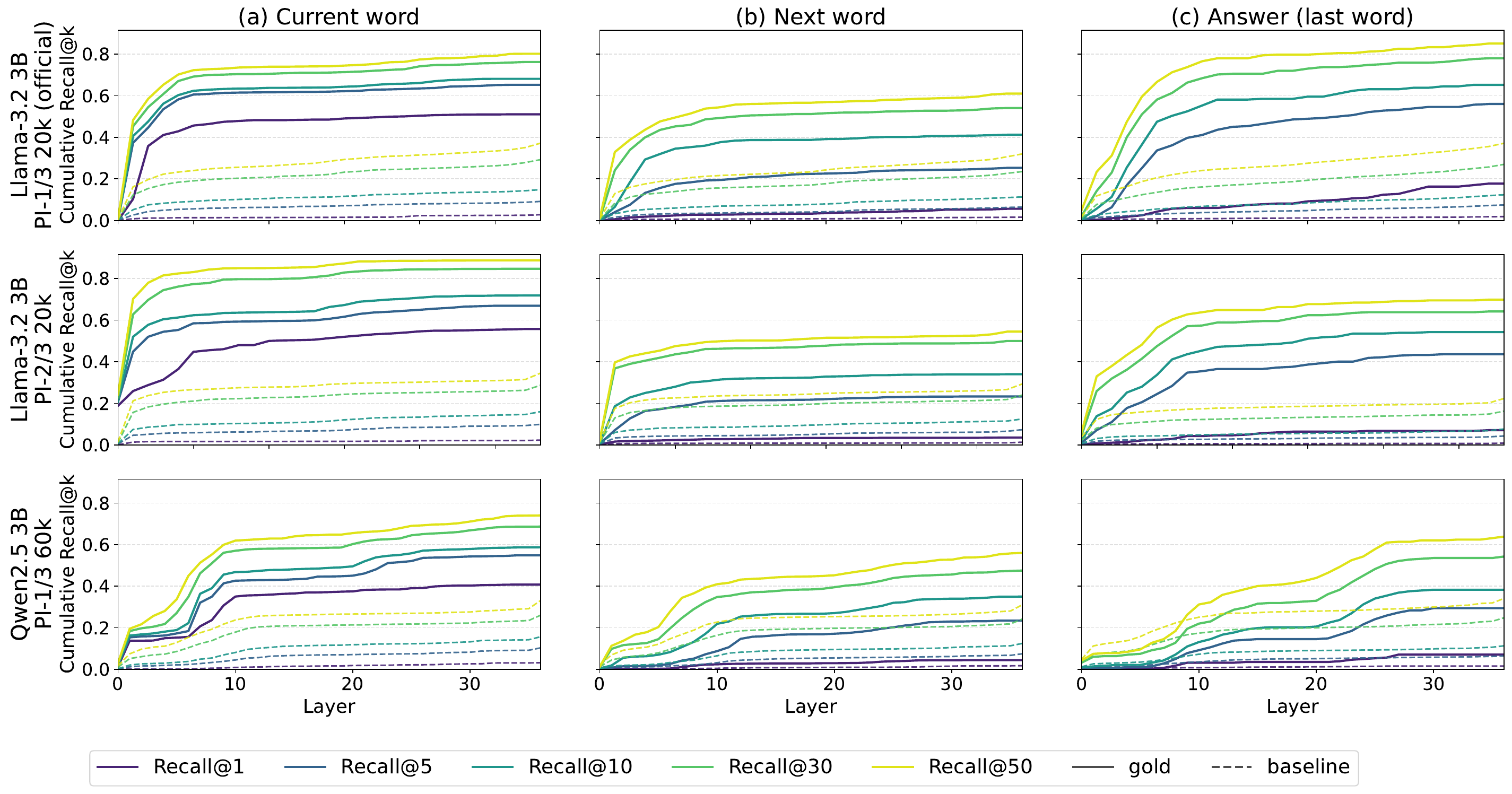}
    \caption{
\textbf{Full cumulative recovery using the latent lens.}
Cumulative Recall@\(k\) across layers for the current word, next word,
and final answer, obtained by retrieving the nearest contextualized
text representations. Solid curves show recovery of the gold target,
while dashed curves show the matched-frequency control. Speech hidden
states are strongly aligned with the corresponding textual
representations, particularly for the current spoken word.
}
\label{fig:transcription_latent}
\end{figure*}

Figure~\ref{fig:transcription_logit_lens} reports the full cumulative
Recall@\(k\) curves for the logit lens in the main text, while
Figures~\ref{fig:transcription_latent} and
\ref{fig:transcription_recall_all_by_layer_latent} report the
corresponding results for the latent lens and linear probes,
respectively, for \(k\in\{1,5,10,30,50\}\). Across methods and models,
the same overall pattern persists across values of \(k\). Each figure
also reports matched-frequency controls, evaluated with the same
span-level and cumulative protocol. These baselines remain
substantially below gold-word recovery across methods and targets,
showing that the effect is not explained by lexical frequency alone.

\subsubsection{Layer-Wise Recall and Persistence}
\label{app:persistence}

Our main analysis reports cumulative recall, which measures whether a word has become decodable by a given model depth. To identify the layers at which the transcription signal is actively present, we additionally compute non-cumulative, layer-wise recall. 

\begin{figure*}[t]
    \centering
    \includegraphics[width=\textwidth]{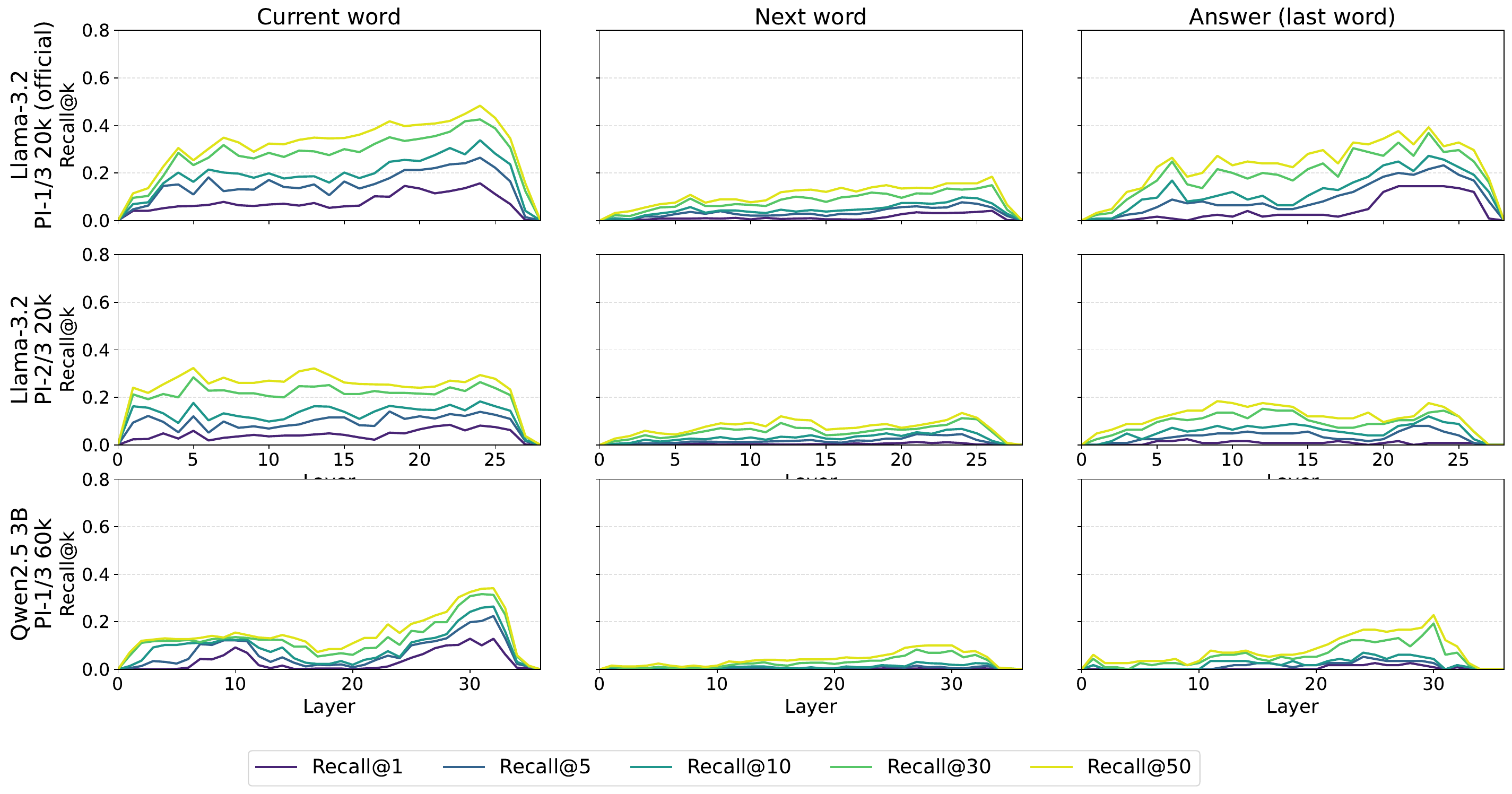}
    \caption{\textbf{Layer-wise recovery using the logit lens.} Layer-wise Recall@(k), computed using the logit lens on speech-token hidden states, is shown for the current transcription word, the next word, and the final answer. Although the models are not explicitly trained for transcription, the current word is reliably recoverable from intermediate layers across models, while next-word and answer-word predictions are weaker but remain decodable. Transcription emerges primarily in the middle layers.}
\label{fig:transcription_recall_all_by_layer}
\end{figure*}

\begin{figure*}[t]
    \centering
    \includegraphics[width=\textwidth]{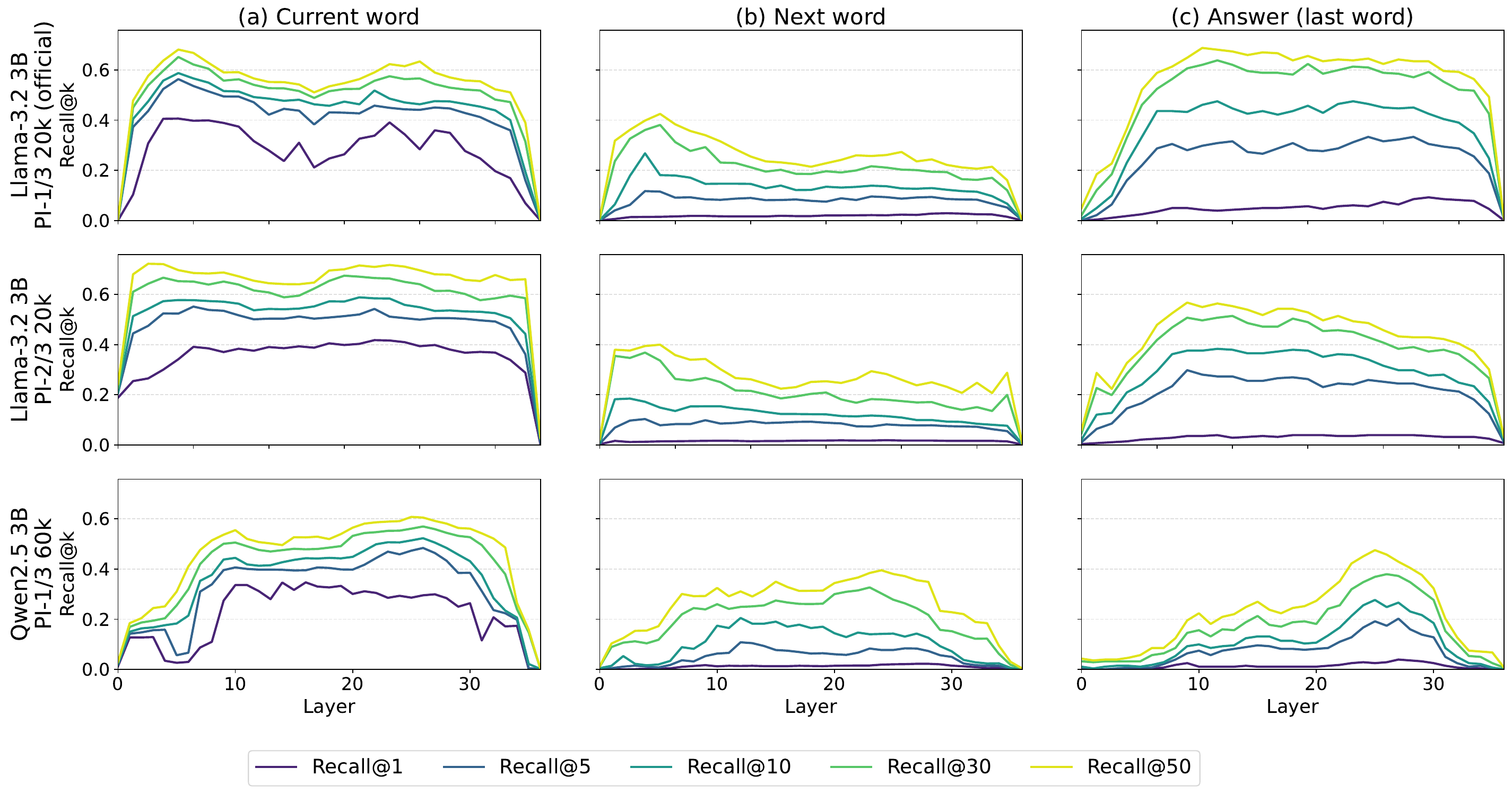}
    \caption{\textbf{Layer-wise recovery using the latent lens.} Layer-wise Recall@(k), computed using the latent lens on speech-token hidden states, is shown for the current transcription word, the next word, and the final answer. Although the models are not explicitly trained for transcription, the current word is reliably recoverable from intermediate layers across models, while next-word and answer-word predictions are weaker but remain decodable.
    Transcription emerges primarily in the middle layers.}
\label{fig:transcription_recall_all_by_layer_latent}
\end{figure*}

\begin{figure*}[t]
    \centering
    \includegraphics[width=\textwidth]{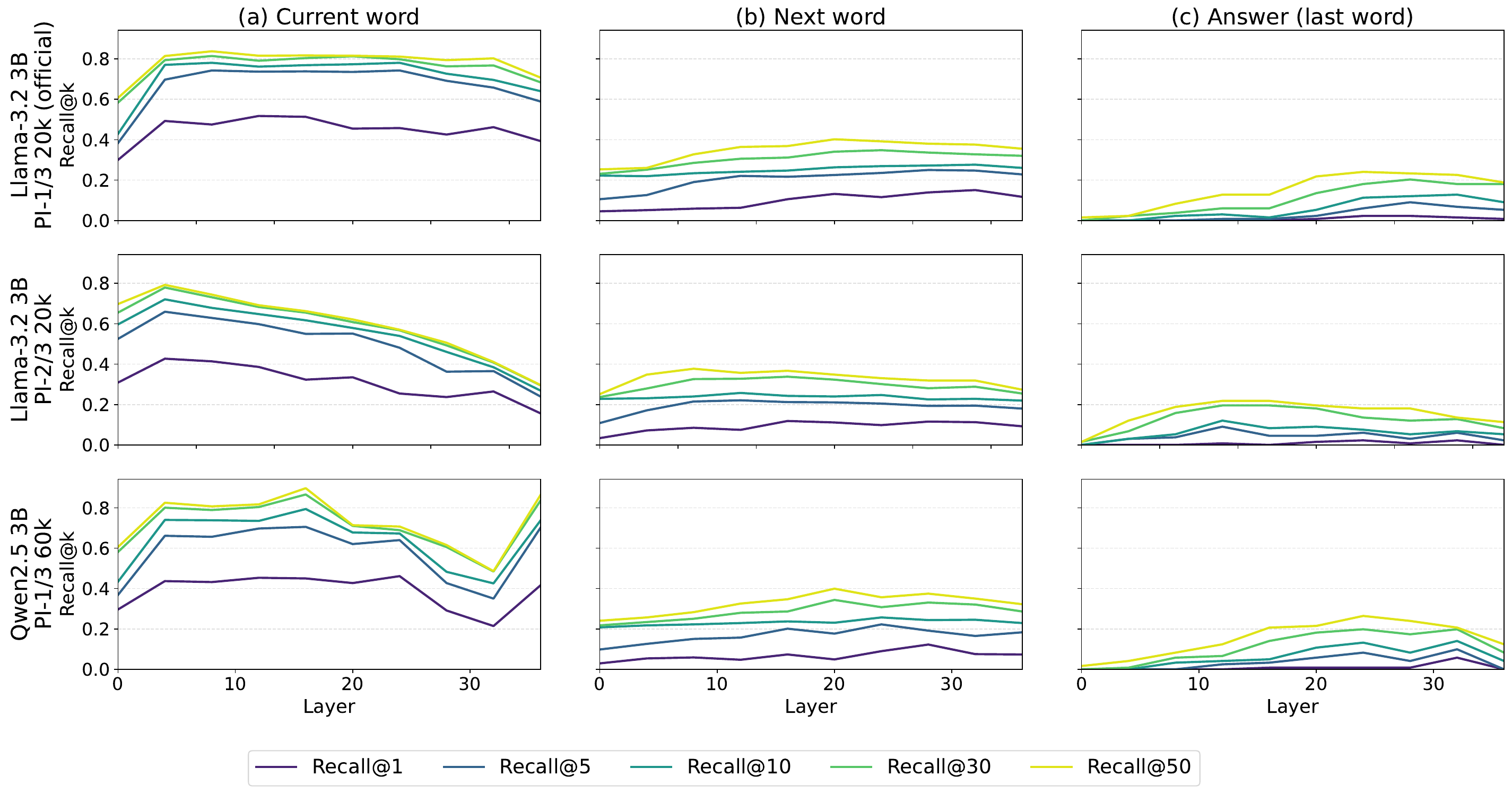}
    \caption{\textbf{Layer-wise recovery using the linear probe.} Layer-wise Recall@(k), computed using the linear probe on speech-token hidden states, is shown for the current transcription word, the next word, and the final answer. Although the models are not explicitly trained for transcription, the current word is reliably recoverable from intermediate layers across models, while next-word and answer-word predictions are weaker but remain decodable. Transcription emerges primarily in the middle layers.}
\label{fig:transcription_recall_all_by_layer_probe}
\end{figure*}

Figures~\ref{fig:transcription_recall_all_by_layer} - \ref{fig:transcription_recall_all_by_layer_probe} present layer-wise recall. Across models, recall generally increases through the intermediate layers, peaks in the late-intermediate layers, and drops sharply near the output. Thus, the transcription signal is not uniformly distributed across model depth, but is most strongly decodable from intermediate hidden states.

We next test whether recovery is restricted to isolated layers or persists across model depth. For each word recovered at least once, we identify its first recovery layer and measure the number of consecutive layers for which the target remains among the top-$k$ predictions at some position in the aligned speech span.

\begin{table}[t]
\centering
\small
\resizebox{\columnwidth}{!}{
\begin{tabular}{lcccc}
\toprule
& \textbf{$\geq$5 layers}
& \textbf{$\geq$8 layers}
& \textbf{$\geq$10 layers}
& \textbf{$\geq$15 layers} \\
\midrule
Recall@1  & 24.8\% & 12.4\% & 7.4\%  & 5.4\% \\
Recall@5  & 31.8\% & 18.7\% & 12.1\% & 7.9\% \\
Recall@10 & 37.7\% & 24.4\% & 17.3\% & 11.1\% \\
\bottomrule
\end{tabular}
}
\caption{Persistence of implicit transcription in Llama-3.2 3B PI-1/3 20k (official). Values report the fraction of word instances recovered at least once that remain recoverable for at least the specified number of layers.}
\label{tab:transcription_persistence}
\end{table}

The persistence results are shown in Table~\ref{tab:transcription_persistence}. At $k=10$, $37.7$\% of the recovered word instances remain decodable for at least $5$ layers, $24.4$\% for at least $8$ layers, and $17.3$\% for at least $10$ layers. Even under the stricter Recall@1 criterion, $24.8$\% of recovered words persist for at least $5$ layers.

These results show that implicit transcription is not limited to isolated layer-wise hits. Although persistence varies across words, a substantial subset remains decodable across several consecutive intermediate layers, supporting the interpretation of an extended intermediate transcription regime rather than a purely transient effect.

\subsubsection{Transcription Examples}
\label{app:trans_ex}
\begin{figure*}[t]
    \centering
    \includegraphics[width=\textwidth]{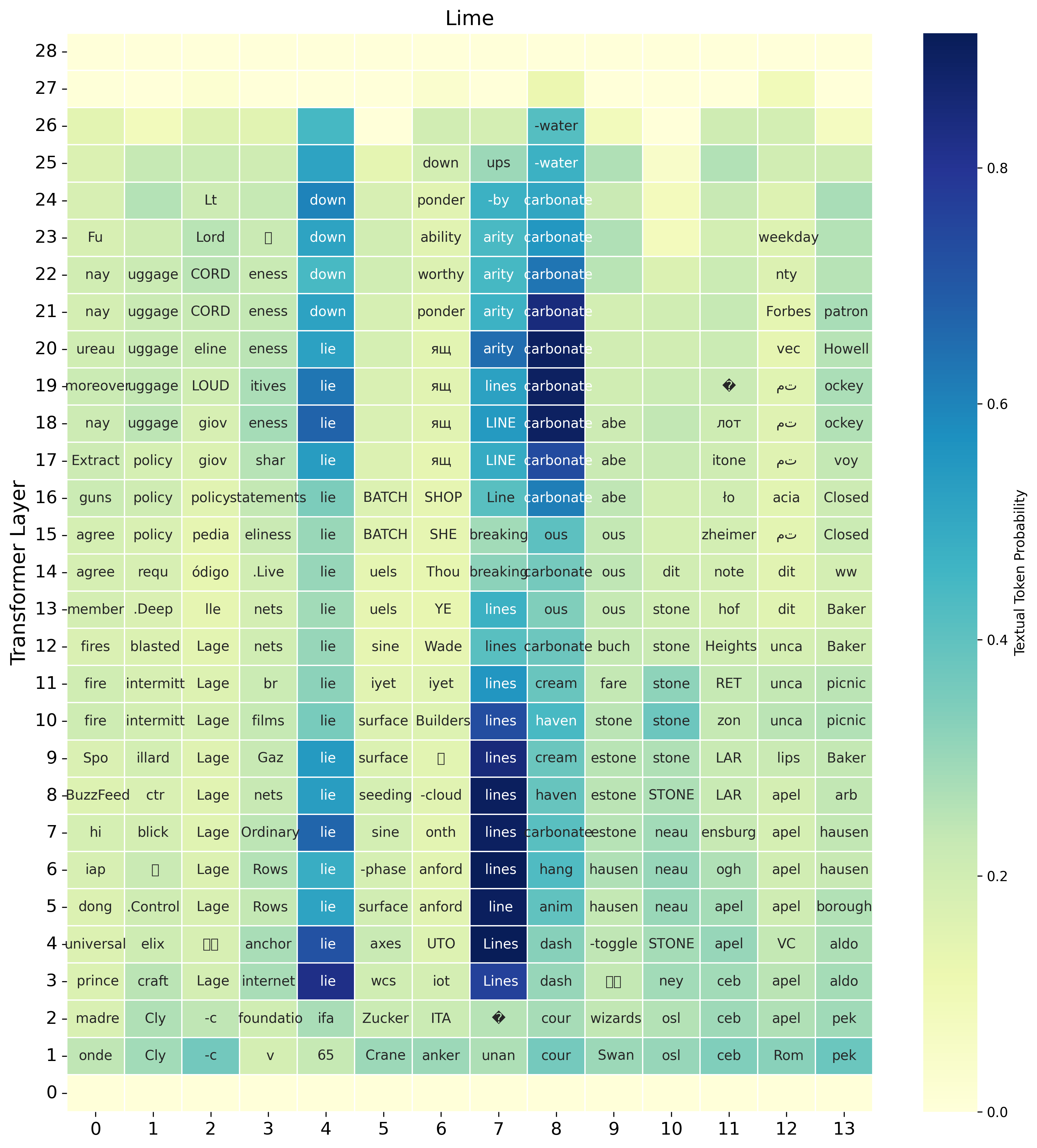}
    \caption{Logit lens of intermediate states for the spoken input "lime", using the Llama-3.2 PI-1/3 (official) model. The example shows both gradual transcription, with an early prediction of 
    "lie", and a transcription error, where the model predicts "line" instead of "lime".}

\label{fig:transcription_lime}
\end{figure*}

\begin{figure*}[t]
    \centering
    \includegraphics[width=\textwidth]{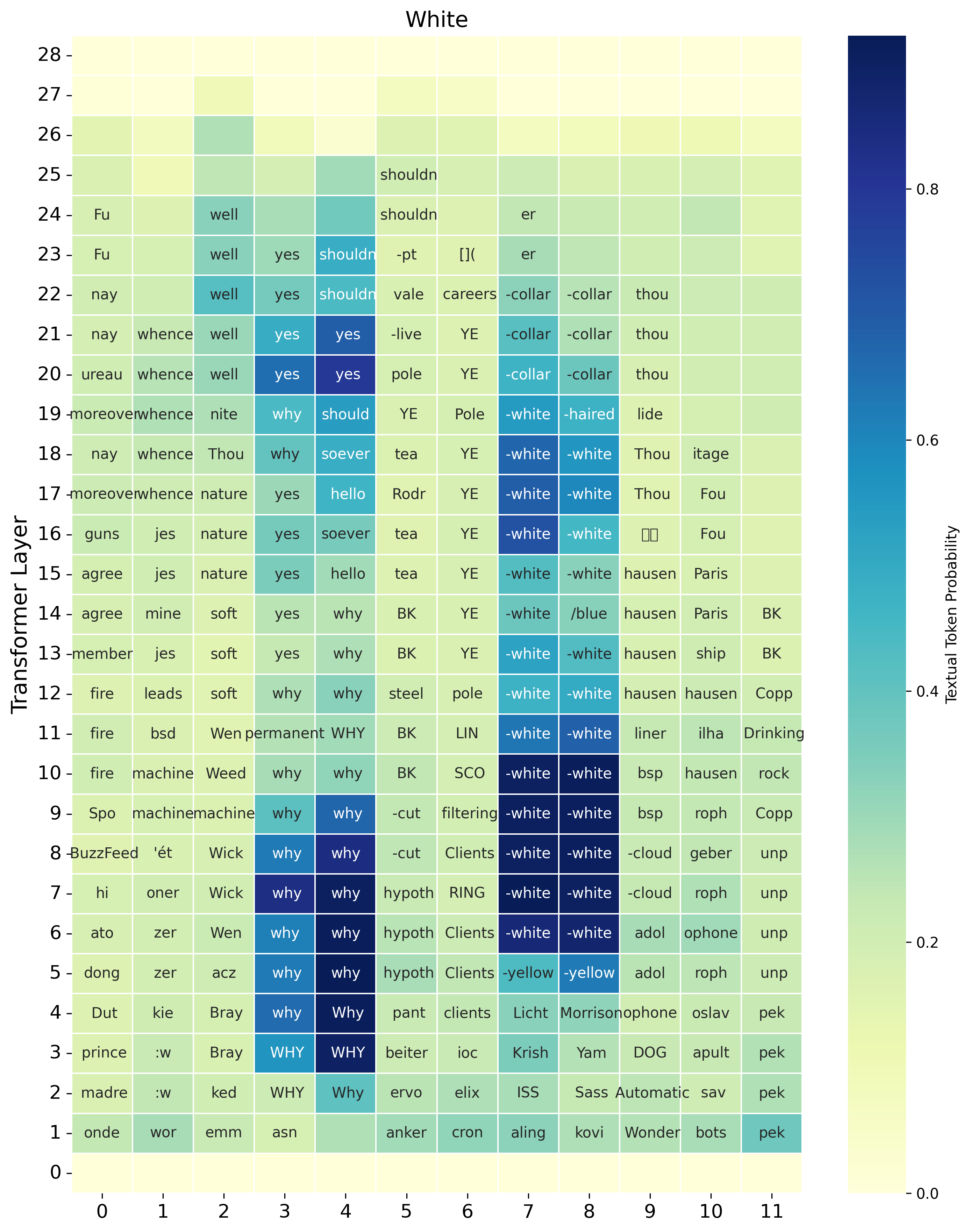}
    \caption{Logit lens of intermediate states for the spoken input "white", using the Llama-3.2 PI-1/3 (official) model. The example shows gradual transcription, where the model first predicts the partial form "why" before later converging to the full word "white".
}

\label{fig:transcription_white}
\end{figure*}

\begin{figure*}[t]
    \centering
    \includegraphics[width=\textwidth]{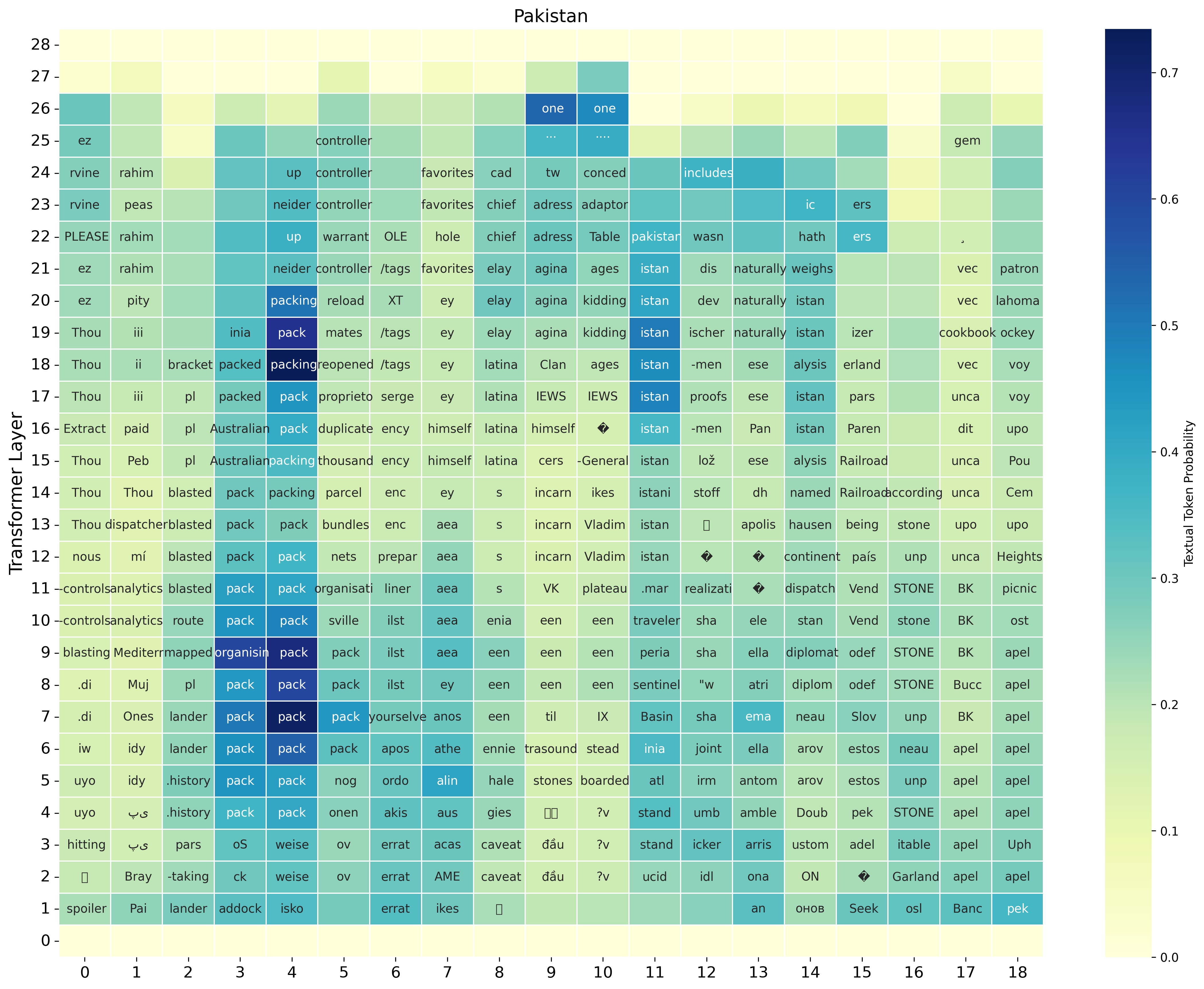}
    \caption{Logit lens of intermediate states for the spoken input "Pakistan", using the Llama-3.2 PI-1/3 (official) model. The example shows gradual transcription, where the model first predicts the partial form "pack'' and later updates toward later components of the word, such as "istan", before converging to "Pakistan".}
\label{fig:transcription_pakistan}
\end{figure*}

\begin{figure*}[t]
    \centering
    \includegraphics[width=\textwidth]{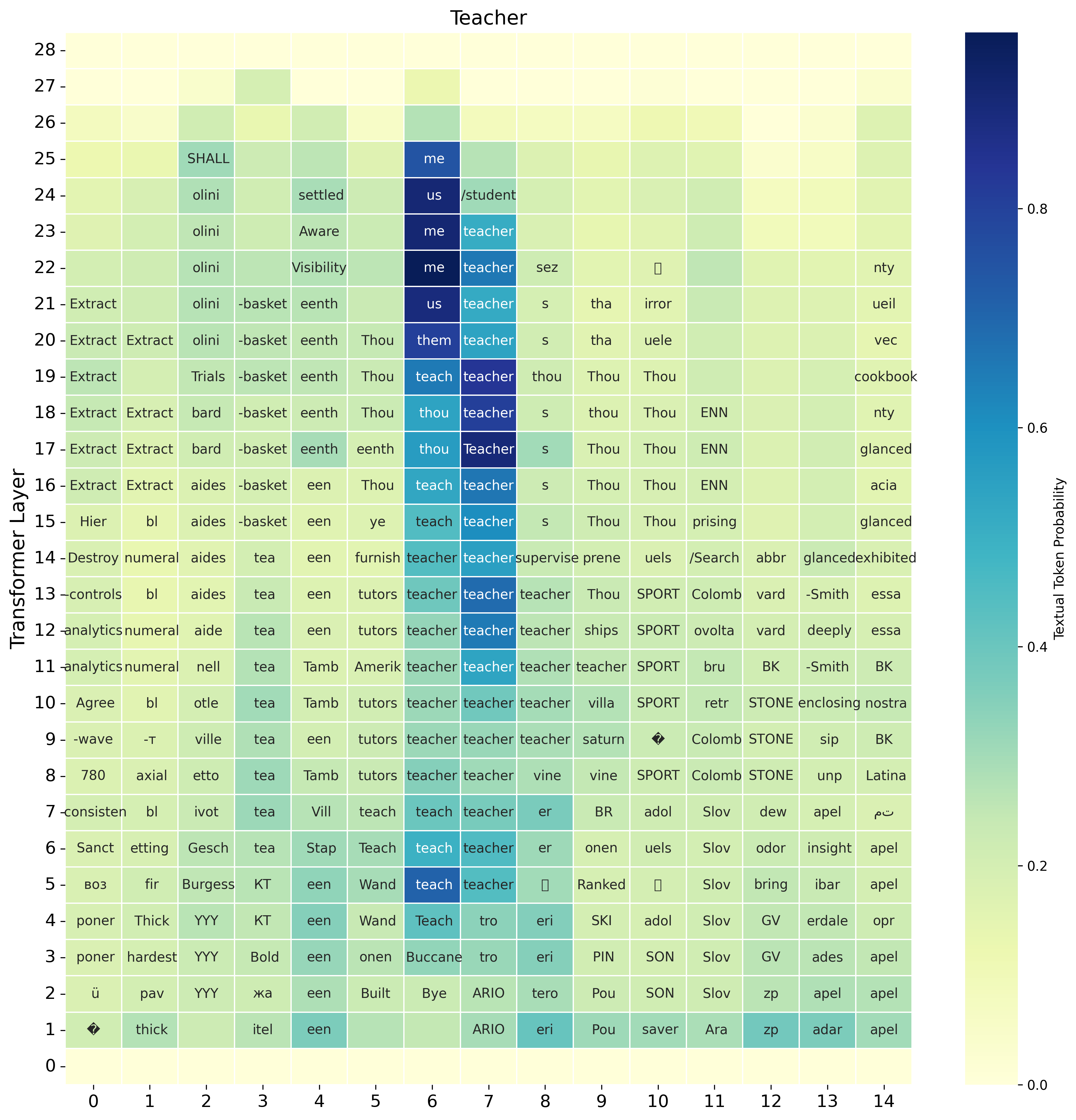}
    \caption{Logit lens of intermediate states for the spoken input "teacher", using the Llama-3.2 PI-1/3 (official) model. The example shows gradual transcription, where the model first predicts "tea", then "teach", and finally converges to the full word "teacher".
}

\label{fig:transcription_teacher}
\end{figure*}

\begin{figure*}[t]
    \centering
    \includegraphics[width=\textwidth]{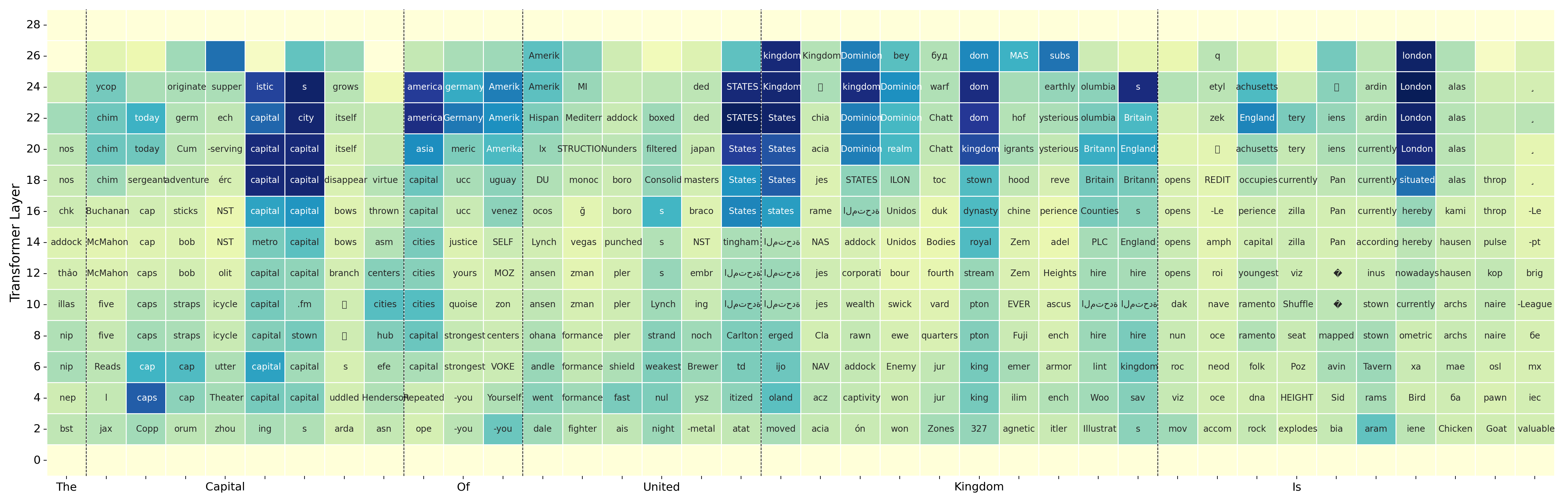}
    \caption{Logit Lens of inner states of the spoken input: "The capital of United Kingdom is...", using Llama-3.2 PI-1/3 (official).This version contains less filters on the time and layer domain and thus show higher resolution in the inner mechanism by logit lens perspective}

\label{fig:transcription_uk_full}
\end{figure*}

\begin{figure*}[t]
    \centering
    \includegraphics[width=\textwidth]{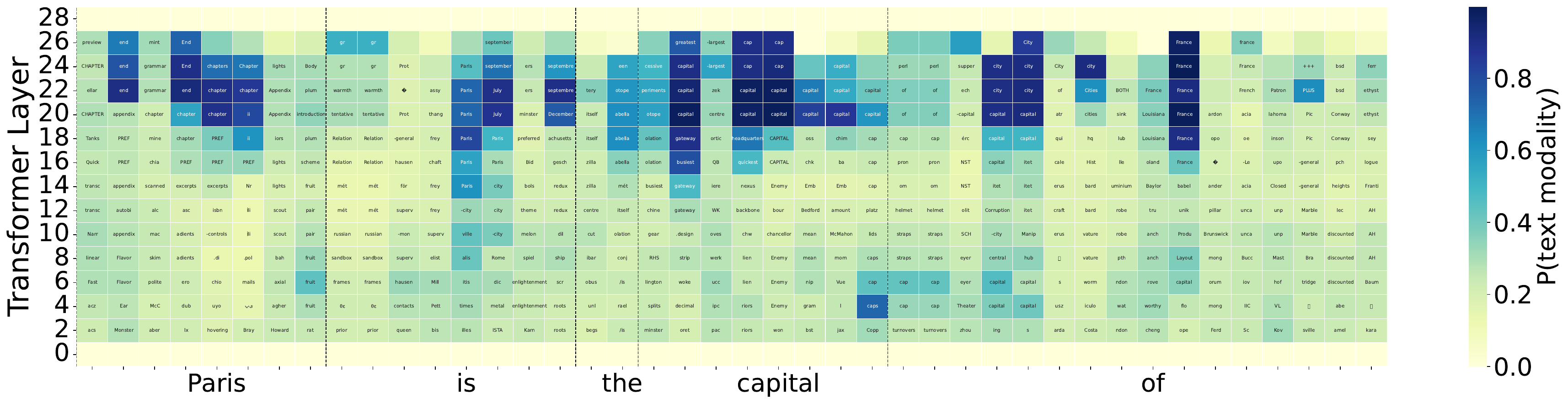}
    \caption{Logit lens of inner states of the spoken input: "Paris is the capital of ...", using Llama-3.2 PI-1/3 (official). }

\label{fig:transcription_paris_sims}
\end{figure*}

\begin{figure*}[t]
    \centering
    \includegraphics[width=\textwidth]{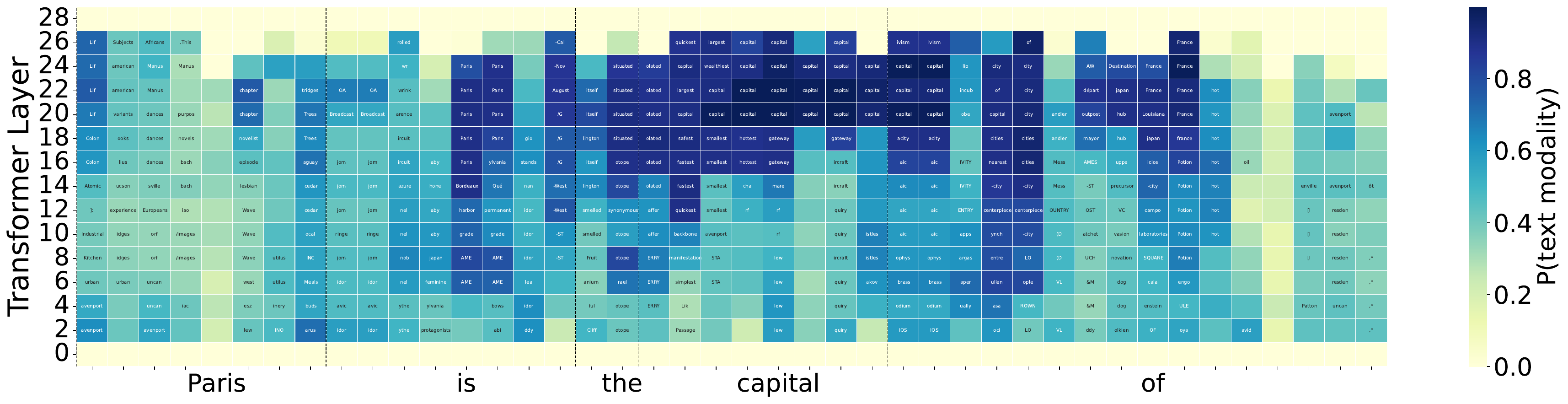}
    \caption{Logit Lens of inner states of the spoken input: "Paris is the capital of ...", using Llama-3.2 PI-1/3 (ours).}

\label{fig:transcription_paris_sims_my}
\end{figure*}

\begin{figure*}[t]
    \centering
    \includegraphics[width=\textwidth]{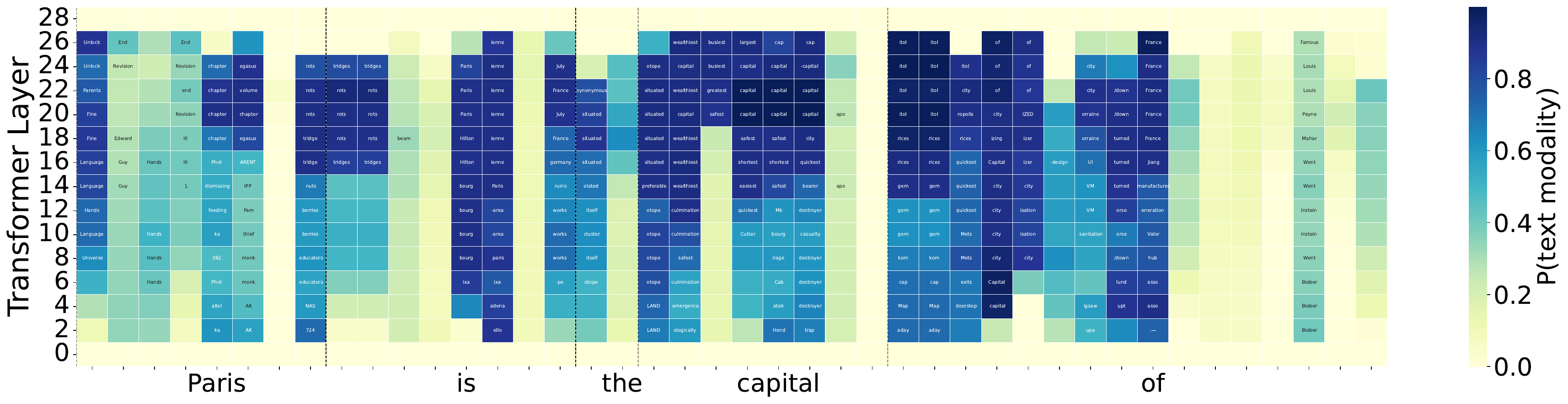}
    \caption{Logit Lens of inner states of the spoken input: "Paris is the capital of ...", using Llama-3.2 PI-2/3.}

\label{fig:transcription_paris_sims07}
\end{figure*}

\begin{figure*}[t]
    \centering
    \includegraphics[width=\textwidth]{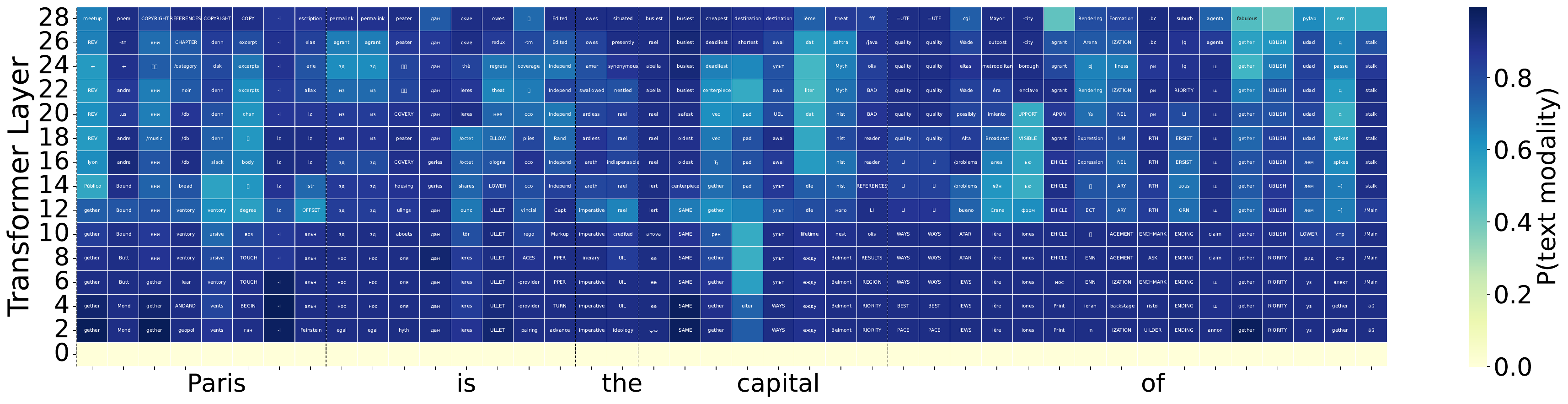}
    \caption{Logit Lens of inner states of the spoken input: "Paris is the capital of...", using Qwen2.5-3B PI-1/3 60k}

\label{fig:transcription_paris_qwen}
\end{figure*}

\begin{figure*}[t]
    \centering
    \includegraphics[width=\textwidth]{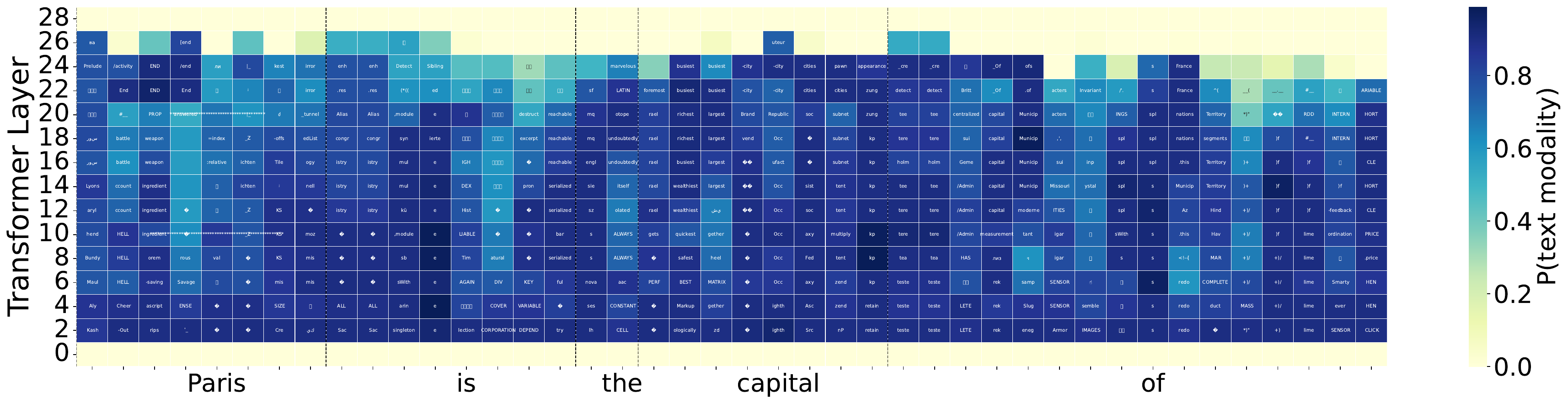}
    \caption{Logit Lens of inner states of the spoken input: "Paris is the capital of...", using Qwen2.5-1.5B PI-1/3 42K}

\label{fig:transcription_paris_qwen_2}
\end{figure*}

\begin{figure*}[t]
    \centering
    \includegraphics[width=\textwidth]{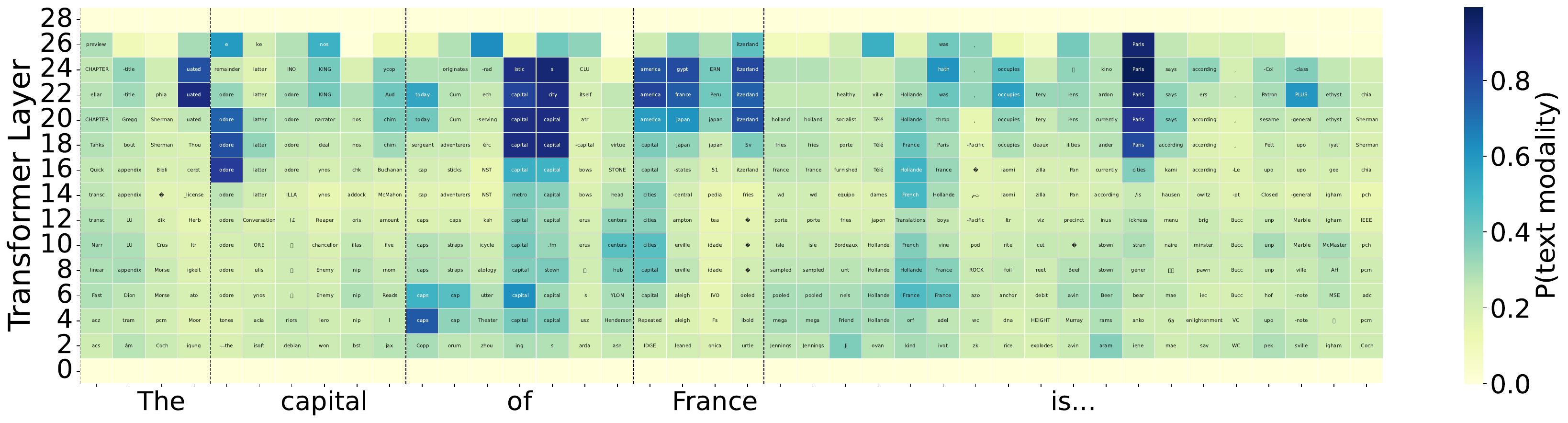}
    \caption{Logit Lens of inner states of the spoken input: "The capital of France is...", using Llama-3.2 PI-1/3 (official). }

\label{fig:transcription_france_sims}
\end{figure*}

\begin{figure*}[t]
    \centering
    \includegraphics[width=\textwidth]{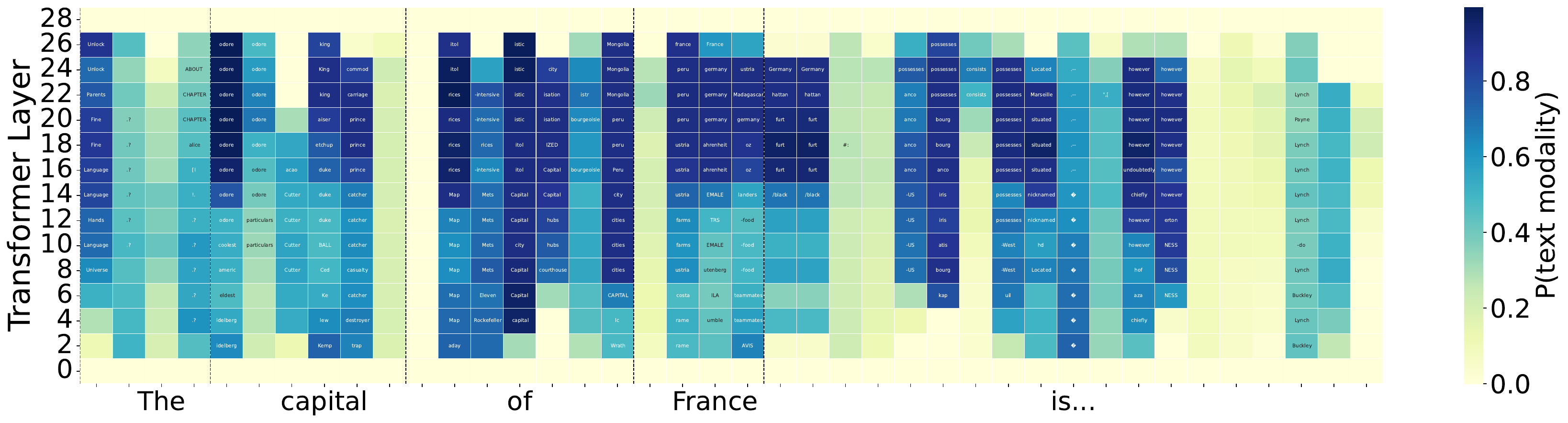}
    \caption{Logit Lens of inner states of the spoken input: "The capital of France is...", using Llama-3.2 PI-2/3}

\label{fig:transcription_france_sims07}
\end{figure*}

\begin{figure*}[t]
    \centering
    \includegraphics[width=\textwidth]{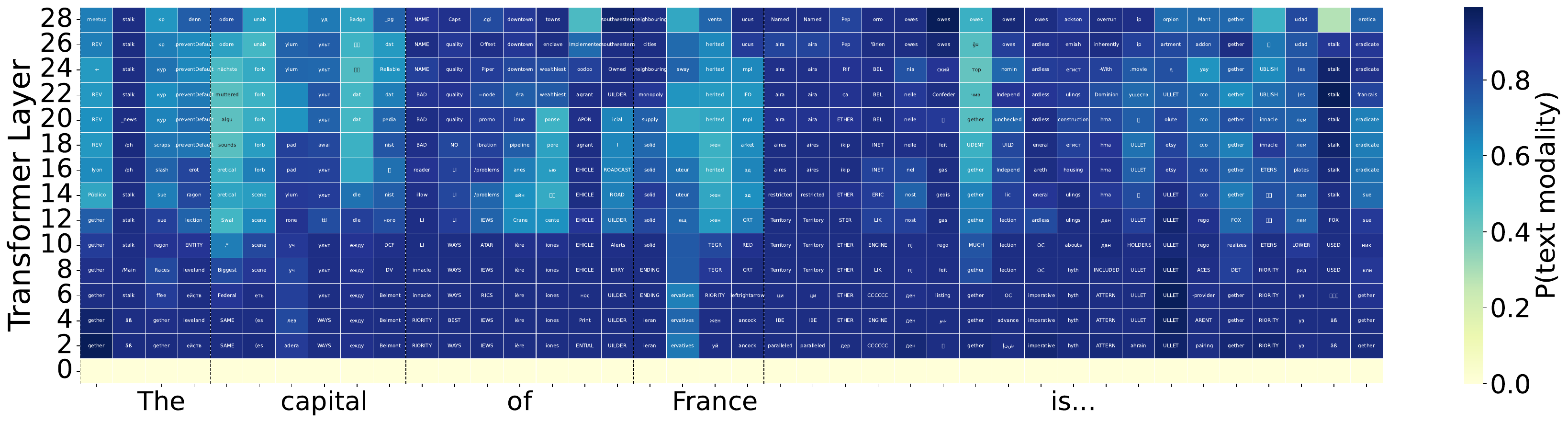}
    \caption{Logit Lens of inner states of the spoken input: "The capital of France is...", using Qwen2.5-3B PI-1/3 60k. }

\label{fig:transcription_france_rand}
\end{figure*}

\begin{figure*}[t]
    \centering
    \includegraphics[width=\textwidth]{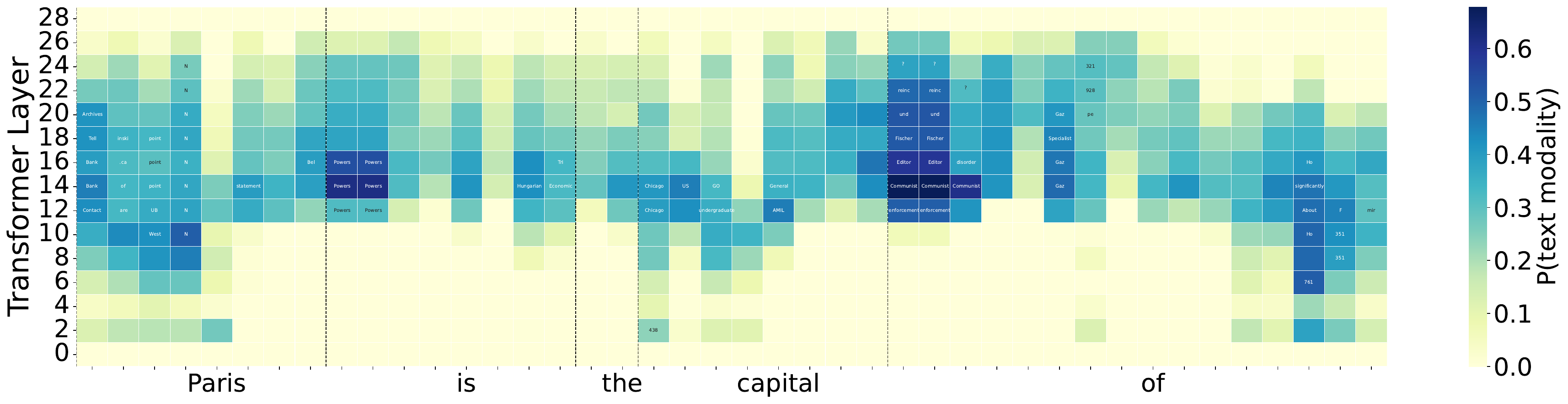}
    \caption{Logit Lens of inner states of the spoken input: "Paris is the capital of ...", using Llama3.2-3B RST}

\label{fig:sims_rand_mix}
\end{figure*}

\begin{figure*}[t]
    \centering
    \includegraphics[width=\textwidth]{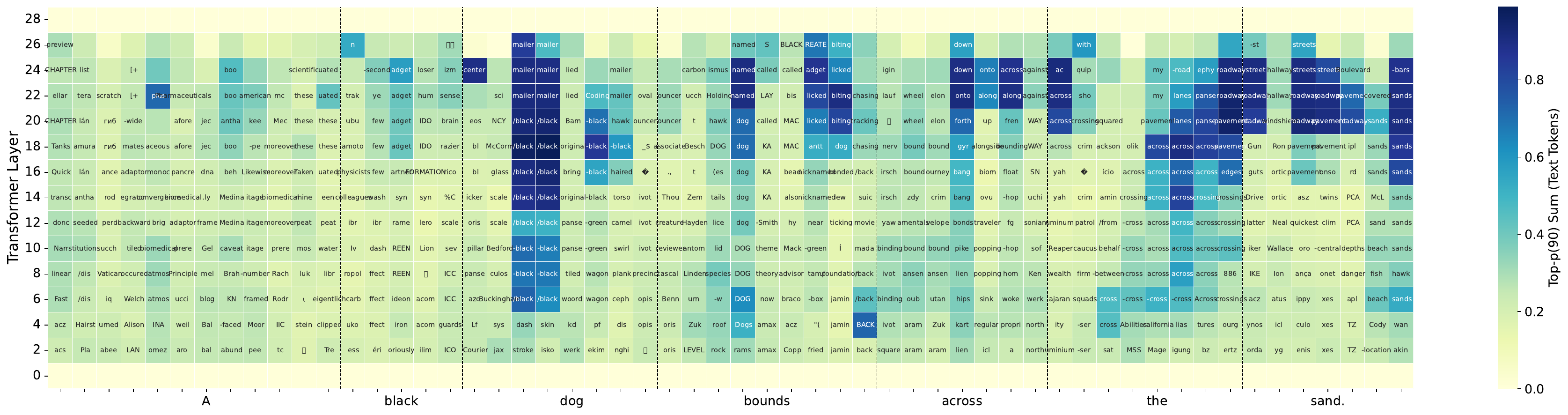}
    \caption{Logit Lens of inner states of the natural spoken input from Flickr8k dataset: "A black dog bounds across the sannd.", using using Llama-3.2 PI-1/3 (official)}

\label{fig:sims_natural_speech_1}
\end{figure*}

\begin{figure*}[t]
    \centering
    \includegraphics[width=\textwidth]{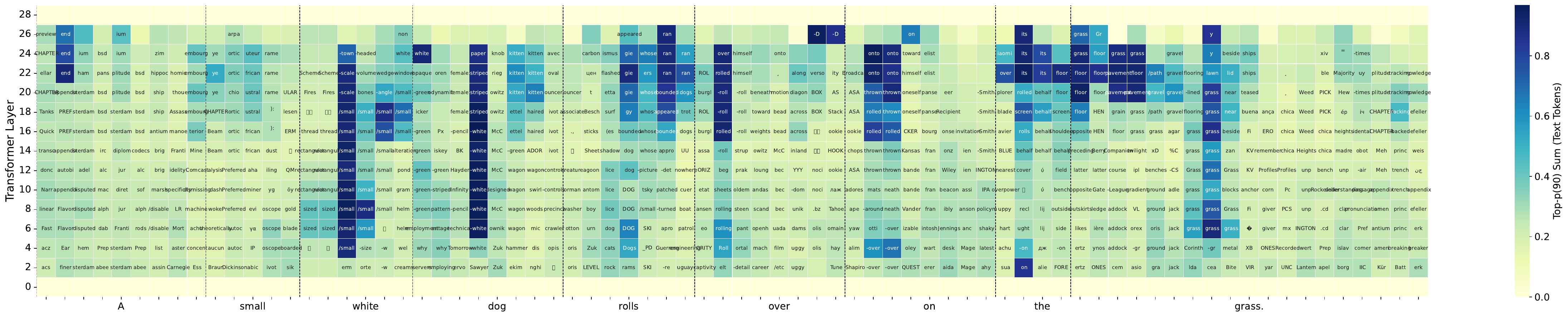}
    \caption{Logit Lens of inner states of the natural spoken input from Flickr8k dataset: "A small white dog rolls over on the grass.", using using Llama-3.2 PI-1/3 (official)}

\label{fig:sims_natural_speech_2}
\end{figure*}

Figures~\ref{fig:transcription_lime}-\ref{fig:sims_natural_speech_2} show logit lens analyses across different models and spoken inputs. Overall, the trends are similar to those presented in the main paper: several models exhibit clear implicit transcription and next word prediction. However, the strength and extent of this phenomenon vary across models.

\begin{figure*}[t]
    \centering
    \includegraphics[width=\textwidth]{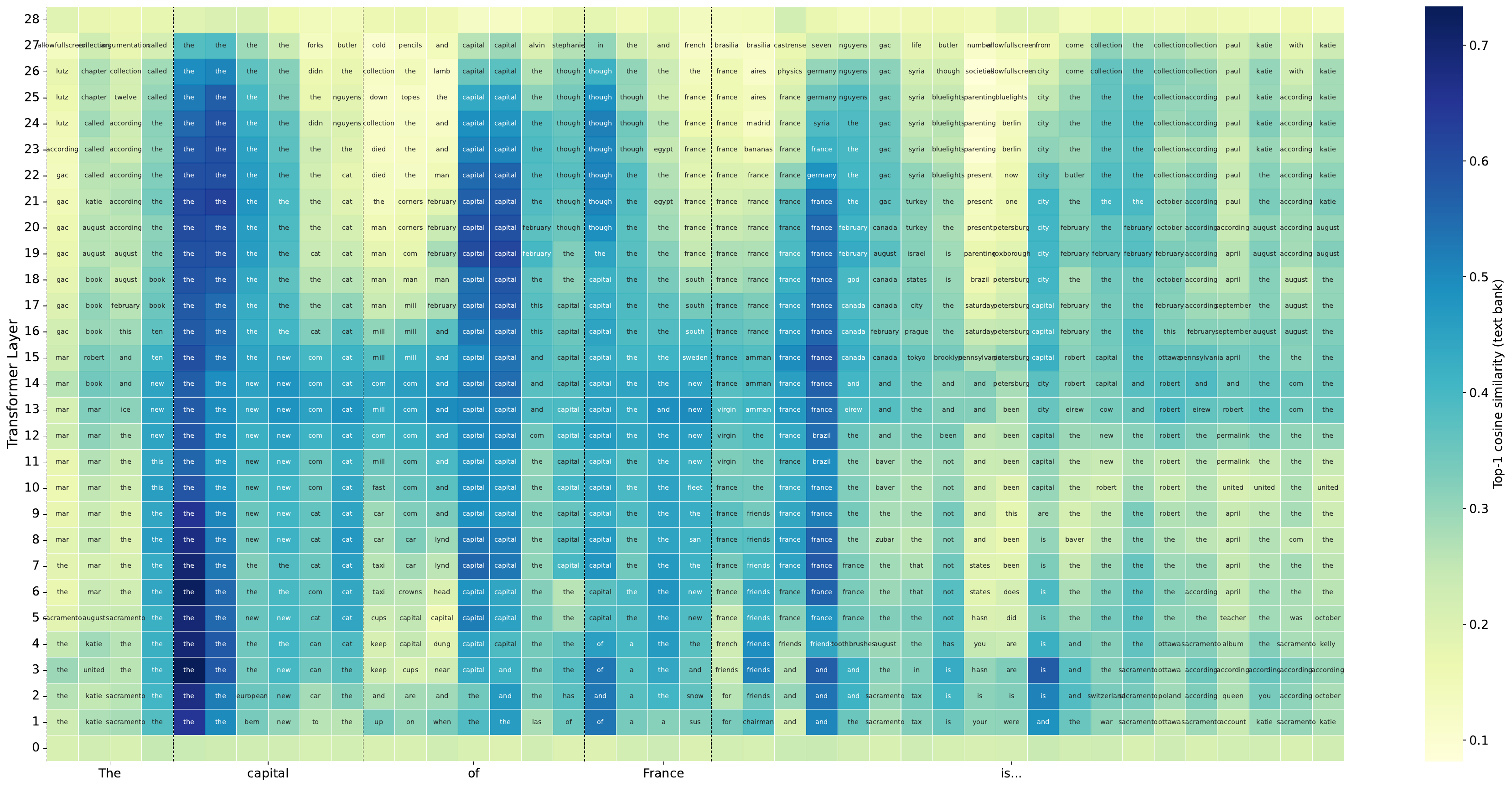}
    \caption{Latent Lens of inner states of the spoken input: "The capital of France is...", using Llama-3.2 PI-1/3 (official). }

\label{fig:transcription_france_sims_latent}
\end{figure*}

Figures~\ref{fig:transcription_lime}-\ref{fig:transcription_teacher} show examples of gradual transcription and possible transcription errors for single-word inputs. Figures~\ref{fig:transcription_uk_full}-\ref{fig:transcription_france_rand} show examples of transcription for different inputs across different models, while Figure~\ref{fig:sims_rand_mix} shows a case in which transcription does not occur at all, for a randomly initialized model. This is consistent with the low transcription capacity observed for this model in Table~\ref{tab:cur_next_ans_results_third} and Section~\ref{sec:res}.
Natural Speech implicit transcription examples can be found in figures 
Figure \ref{fig:sims_natural_speech_1} - \ref{fig:sims_natural_speech_2}, and Figure \ref{fig:transcription_france_sims_latent} contains an example for implicit transcription given by Latent Lens.
In all figures, cells are colored by textual-token probability, dark blue indicates high probability and light yellow is zero textual probability. Transcriptions are written for textual token according to the most probable token given by the decoding method.

\subsubsection{Transcription Recall For Different Top Ks Across Different Models}
\label{app:trans_recall}
\begin{figure*}[t]
    \centering
    \includegraphics[width=\textwidth]{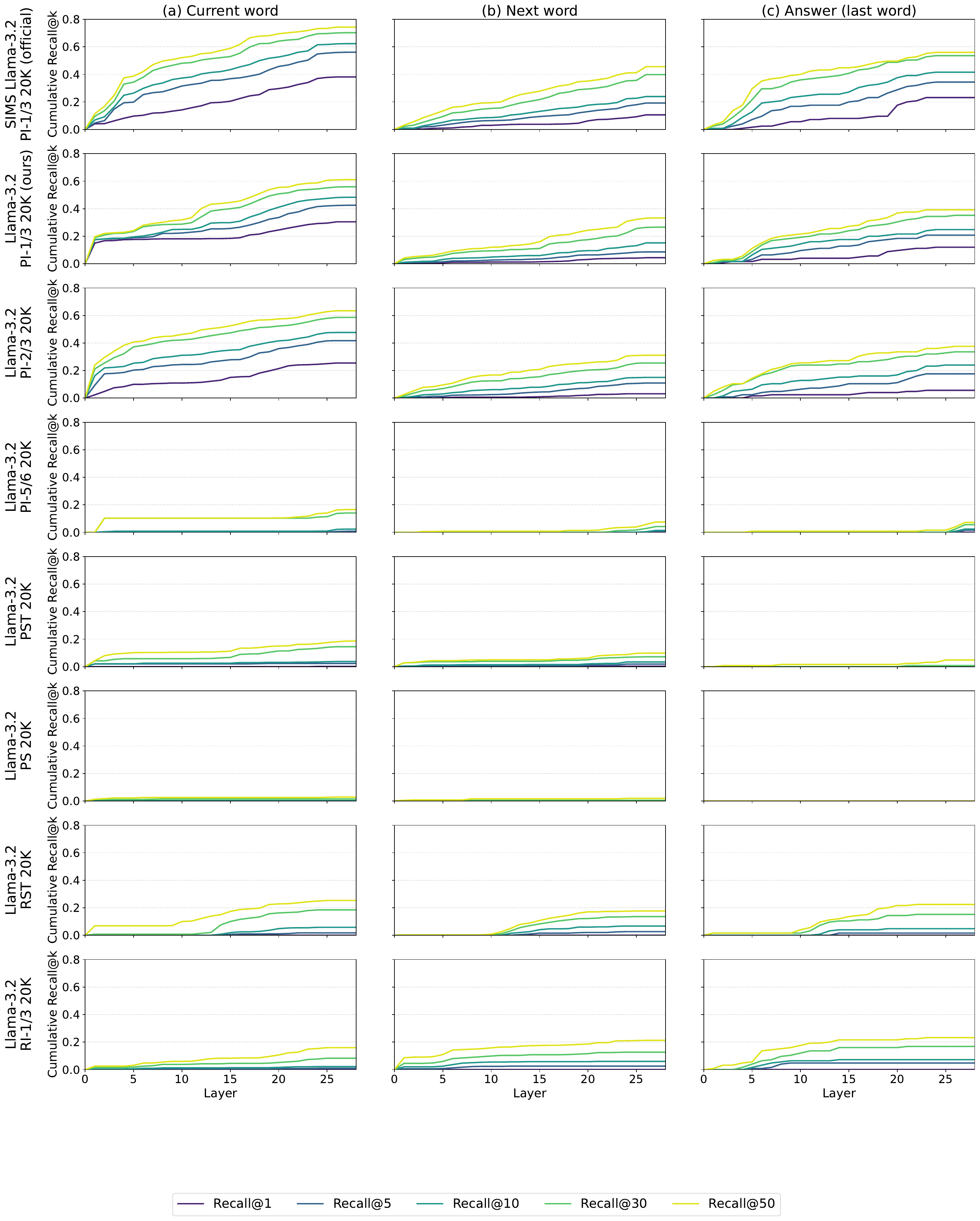}
    \caption{\textbf{Implicit transcription and textual continuation emerge in speech hidden states for text pre-trained with interleaving.} We apply the logit lens to speech-token hidden states and report Recall@k up to a given layer, for the current transcription word, the next word, and the final answer. Although the models are not explicitly trained for transcription, current-word
    transcription emerges reliably in intermediate layers across models, while next-word and answer-word predictions
    are weaker but still decodable. }

\label{fig:transcription_recall_all_k_sims}
\end{figure*}

We analyze $\mathrm{Recall@}k$ for the Llama3.2-3B variants with $k \in \{1,5,30\}$. The results show consistent trends across different values of $k$ and are reported in Tables~\ref{tab:cur_next_ans_results_k1}, \ref{tab:cur_next_ans_results_k5}, and~\ref{tab:cur_next_ans_results_k30}.
The cumulative recall patterns across all values of $k$ and all models are shown in Figure~\ref{fig:transcription_recall_all_k_sims}.

\subsubsection{Natural-Speech Results by Corpus}
\label{app:natural_speech}

We evaluate 300 utterances from each of seven natural-speech corpora:
People's Speech, LibriSpeech, VoxPopuli, TED-LIUM, the Spoken
Wikipedia Corpus (SWC), VCTK, and Flickr8k \citep{harwath2015deepmultimodalsemanticembeddings}. These corpora span
different speakers, recording conditions, and speaking styles.

Tables~\ref{tab:natural_latent}--\ref{tab:natural_probe} report
corpus-level cumulative Recall for the logit lens, latent lens, and
linear probes, respectively. Each entry reports gold-word recovery,
with the matched-frequency baseline in parentheses. Across corpora,
methods, and models, current-word recovery is consistently stronger
than next-word recovery, and both generally remain substantially above
their frequency-matched controls. Absolute performance varies across
corpora, indicating sensitivity to their acoustic and linguistic
properties, but the transcription signal is not restricted to a
particular speech domain.

Natural-speech implicit transcription examples are shown in
Figures~\ref{fig:sims_natural_speech_1}--\ref{fig:sims_natural_speech_2}.

\subsection{Model Evaluation}
\label{app:model_eval}
\begin{figure*}[h]
    \centering
    \includegraphics[width=1\linewidth]
{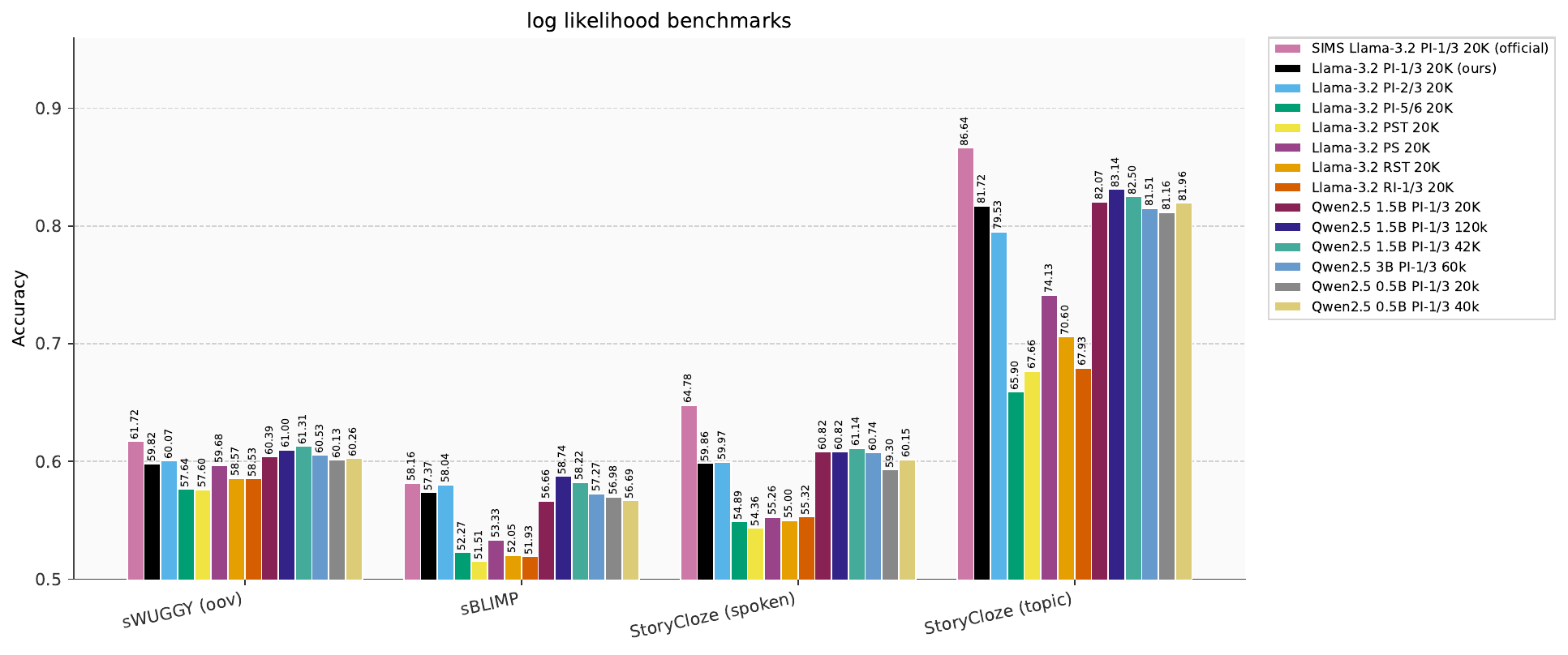}
    \caption{Log Likelihood based evaluations for all models we used}

\label{fig:models_eval}
\end{figure*}

For reference, and as a proxy for the general semantic abilities of the trained models we also report standard likelihood speech metrics in Figure~\ref{fig:models_eval}.
We note that indeed all models perform as expected with models randomly initialized and without interleaving under-performing those with. We also see that our version of SIMS performs comparably to the official version once again showing that our training setup works as expected.

\subsection{AI Tools Usage}
AI tools have been used to assist in fixing grammar mistakes and sentence paraphrasing. Additionally, AI tools have been partially used to enhance code implementations. However, the authors carefully reviewed all content, ensuring these tools were only used as supportive aids and in a responsible manner.

\begin{table*}[t]
\centering
\small
\begin{tabular}{lrl}
\toprule
\textbf{Category} & \textbf{\# Examples} & \textbf{Example} \\
\midrule
Colors & 16 & \textit{The color of grass is} $\rightarrow$ \textit{green} \\
Days of the week & 7 & \textit{The day that comes after Sunday is} $\rightarrow$ \textit{Monday} \\
Months of the year & 14 & \textit{The month that comes after January is} $\rightarrow$ \textit{February} \\
Object functions & 21 & \textit{People use eyes to} $\rightarrow$ \textit{see} \\
Common-sense facts & 13 & \textit{Water freezes into} $\rightarrow$ \textit{ice} \\
Country languages & 10 & \textit{The official language of France is} $\rightarrow$ \textit{French} \\
Family relations & 18 & \textit{The mother of mother is called} $\rightarrow$ \textit{grandmother} \\
Numerical facts & 27 & \textit{The number of seconds in a minute is} $\rightarrow$ \textit{60} \\
Opposites & 35 & \textit{the opposite of hot is} $\rightarrow$ \textit{cold} \\
Baby animals and professions & 10 & \textit{a baby dog is called a} $\rightarrow$ \textit{puppy} \\
Capital cities & 30 & \textit{The capital of France is} $\rightarrow$ \textit{Paris} \\
Simple arithmetic & 58 & \textit{one plus one equals} $\rightarrow$ \textit{2} \\
Number sequences & 23 & \textit{the number after one is} $\rightarrow$ \textit{two} \\
\midrule
\textbf{Total} & \textbf{282} & -- \\
\bottomrule
\end{tabular}
\caption{Statistics and representative examples from the factual-completion dataset. Each example consists of a short prompt with the final answer omitted and a single target completion.}
\label{tab:dataset_statistics}
\end{table*}

\begin{figure*}[t]
    \centering
    \includegraphics[width=\textwidth]{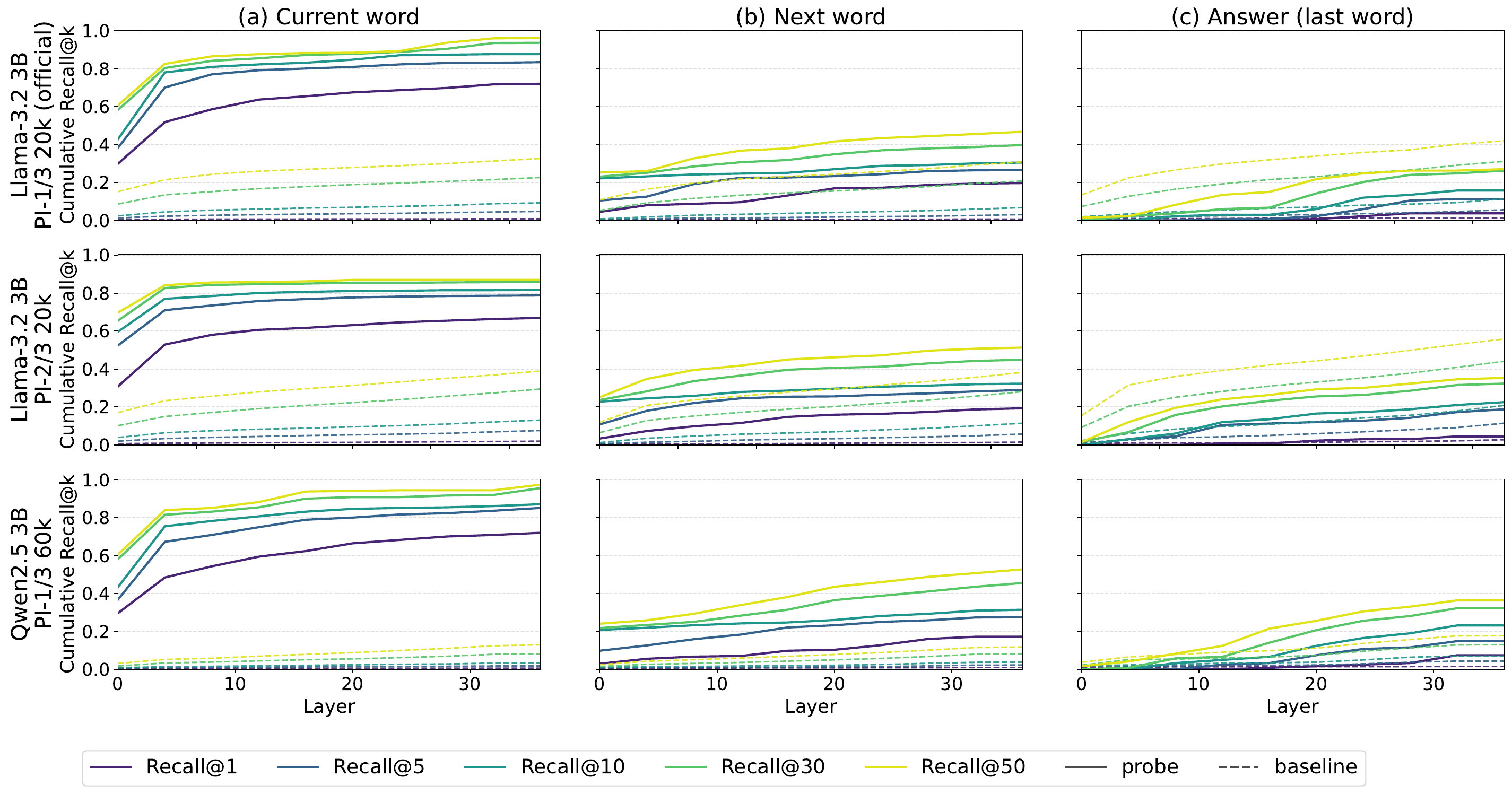}
    \caption{
\textbf{Full cumulative recovery using linear probes.}
Cumulative Recall@\(k\) across layers for the current word, next word, and final answer. Separate supervised linear probes are trained for each model, layer, and prediction target while the Speech LM remains frozen. Solid curves show recovery of the gold target, while dashed
curves show the matched-frequency control. The high recall demonstrates
that lexical and continuation information is strongly linearly
recoverable from intermediate speech hidden states.
}

\label{fig:transcription_probe}
\end{figure*}

\begin{table*}[t]
\centering
\small
\begin{tabular}{l|cccc|ccc}
\toprule
\textbf{Model} 
& \textbf{Text pre-trained}
& \textbf{Text}
& \textbf{Inter.}
& \textbf{Inter. frac.}
& \textbf{Cur}
& \textbf{Next}
& \textbf{Ans} \\
\midrule
SIMS Llama-3.2 PI-1/3 (official) 
& \checkmark & \checkmark & \checkmark & \(1/3\) 
& \textbf{\underline{37.19}} & \textbf{\underline{10.31}} & \textbf{\underline{23.20}} \\
\midrule
Llama-3.2 PI-1/3 (ours) 
& \checkmark & \checkmark & \checkmark & \(1/3\) 
& \textbf{32.66} & \textbf{4.84} & \textbf{12.80} \\

Llama-3.2 PI-2/3
& \checkmark & \checkmark & \checkmark & \(2/3\) 
& \textbf{27.34} & \textbf{3.59} & \textbf{5.60} \\

Llama-3.2 PI-5/6
& \checkmark & \checkmark & \checkmark & \(5/6\) 
& 0.00 & 0.00 & 0.00 \\

Llama-3.2 PST
& \checkmark & \checkmark & x & -- 
& 0.16 & 0.31 & 0.00 \\

Llama-3.2 PS
& \checkmark & x & x & -- 
& 0.00 & 0.00 & 0.00 \\

Llama-3.2 RST
& x & \checkmark & x & -- 
& 0.00 & 0.00 & 0.00 \\

Llama-3.2 RI-1/3
& x & \checkmark & \checkmark & \(1/3\) 
& 0.00 & 0.00 & 0.00 \\
\bottomrule
\end{tabular}
\caption{
Recall@1 for different Speech LMs, i.e., the percentage of examples in which the correct current word (Cur), next word (Next), or answer word (Ans) appears as the top predicted token across the relevant spoken word in any transformer layer. 
}
\label{tab:cur_next_ans_results_k1}
\end{table*}

\begin{table*}[t]
\centering
\small
\begin{tabular}{l|cccc|ccc}
\toprule
\textbf{Model} 
& \textbf{Text pre-trained}
& \textbf{Text}
& \textbf{Inter.}
& \textbf{Inter. frac.}
& \textbf{Cur}
& \textbf{Next}
& \textbf{Ans} \\
\midrule
SIMS Llama-3.2 PI-1/3 (official) 
& \checkmark & \checkmark & \checkmark & \(1/3\) 
& \textbf{\underline{55.16}} & \textbf{\underline{19.22}} & \textbf{\underline{34.40}} \\
\midrule
Llama-3.2 PI-1/3 (ours) 
& \checkmark & \checkmark & \checkmark & \(1/3\) 
& \textbf{42.81} & \textbf{8.91} & \textbf{21.60} \\

Llama-3.2 PI-2/3
& \checkmark & \checkmark & \checkmark & \(2/3\) 
& \textbf{43.75} & \textbf{11.87} & 18.40 \\

Llama-3.2 PI-5/6
& \checkmark & \checkmark & \checkmark & \(5/6\) 
& 1.09 & 1.09 & 3.20 \\

Llama-3.2 PST
& \checkmark & \checkmark & x & -- 
& 2.34 & 1.87 & 0.00 \\

Llama-3.2 PS
& \checkmark & x & x & -- 
& 0.00 & 0.00 & 0.00 \\

Llama-3.2 RST
& x & \checkmark & x & -- 
& 1.72 & 2.66 & 1.60 \\

Llama-3.2 RI-1/3
& x & \checkmark & \checkmark & \(1/3\) 
& 0.94 & 2.34 & 4.80 \\
\bottomrule
\end{tabular}
\caption{
Recall@5 for different Speech LMs, i.e., the percentage of examples in which the correct current word (Cur), next word (Next), or answer word (Ans) appears among the top-5 predicted tokens across the relevant spoken word in any transformer layer, using the logit lens.
}
\label{tab:cur_next_ans_results_k5}
\end{table*}

\begin{table*}[t]
\centering
\small
\begin{tabular}{l|cccc|ccc}
\toprule
\textbf{Model} 
& \textbf{Text pre-trained}
& \textbf{Text}
& \textbf{Inter.}
& \textbf{Inter. frac.}
& \textbf{Cur}
& \textbf{Next}
& \textbf{Ans} \\
\midrule
SIMS Llama-3.2 PI-1/3 (official) 
& \checkmark & \checkmark & \checkmark & \(1/3\) 
& \textbf{\underline{73.75}} & \textbf{\underline{45.78}} & \textbf{\underline{56.80}} \\
\midrule
Llama-3.2 PI-1/3 (ours) 
& \checkmark & \checkmark & \checkmark & \(1/3\) 
& \textbf{61.56} & \textbf{33.59} & \textbf{40.00} \\

Llama-3.2 PI-2/3
& \checkmark & \checkmark & \checkmark & \(2/3\) 
& \textbf{65.00} & \textbf{32.34} & 38.40 \\

Llama-3.2 PI-5/6
& \checkmark & \checkmark & \checkmark & \(5/6\) 
& 17.03 & 7.50 & 8.80 \\

Llama-3.2 PST
& \checkmark & \checkmark & x & -- 
& 18.91 & 9.38 & 4.80 \\

Llama-3.2 PS
& \checkmark & x & x & -- 
& 2.81 & 1.87 & 0.00 \\

Llama-3.2 RST
& x & \checkmark & x & -- 
& 25.31 & 17.50 & 22.40 \\

Llama-3.2 RI-1/3
& x & \checkmark & \checkmark & \(1/3\) 
& 15.62 & 21.41 & 24.00 \\
\bottomrule
\end{tabular}
\caption{
Recall@30 for different Speech LMs, i.e., the percentage of examples in which the correct current word (Cur), next word (Next), or answer word (Ans) appears among the top-30 predicted tokens across the relevant spoken word in any transformer layer, using the logit lens. 
}
\label{tab:cur_next_ans_results_k30}
\end{table*}

\begin{table*}[t]
\centering
\small
\setlength{\tabcolsep}{4pt}
\begin{tabular}{l|ccc|ccc}
\toprule
& \multicolumn{3}{c|}{\textbf{Current word}}
& \multicolumn{3}{c}{\textbf{Next word}} \\
\textbf{Dataset}
& \textbf{Llama PI-1/3}
& \textbf{Llama PI-2/3}
& \textbf{Qwen PI-1/3}
& \textbf{Llama PI-1/3}
& \textbf{Llama PI-2/3}
& \textbf{Qwen PI-1/3} \\
\midrule
Controlled
& 68.1 (14.8) & 71.9 (16.0) & 58.7 (15.7)
& 41.2 (11.3) & 34.0 (12.5) & 35.0 (12.5) \\
\midrule
People's Speech
& 56.8 (5.4) & 64.0 (5.7) & 55.4 (4.7)
& 23.3 (5.5) & 31.6 (5.9) & 32.0 (5.0) \\
LibriSpeech
& 47.1 (7.0) & 58.4 (7.4) & 46.8 (5.8)
& 23.2 (7.0) & 31.4 (7.5) & 32.6 (6.2) \\
VoxPopuli
& 26.3 (6.4) & 38.5 (6.6) & 33.8 (4.8)
& 16.1 (6.4) & 27.1 (6.8) & 27.3 (5.1) \\
TED-LIUM
& 64.1 (8.6) & 72.1 (8.6) & 58.4 (7.4)
& 31.0 (8.5) & 34.8 (8.9) & 35.8 (7.6) \\
SWC
& 24.5 (6.2) & 32.8 (6.4) & 27.7 (5.5)
& 15.1 (6.2) & 21.7 (6.6) & 22.2 (5.8) \\
VCTK
& 76.3 (10.6) & 77.7 (9.4) & 67.5 (10.0)
& 36.4 (9.6) & 28.1 (9.4) & 33.6 (9.0) \\
Flickr8k
& 68.3 (10.8) & 76.6 (10.1) & 61.5 (9.7)
& 25.5 (8.8) & 30.7 (9.4) & 32.0 (9.3) \\
\midrule
Natural speech
& 40.3 (6.3) & 51.7 (6.8) & 43.6 (5.7)
& 20.0 (6.4) & 28.2 (6.9) & 28.9 (5.9) \\
\bottomrule
\end{tabular}
\caption{
\textbf{Natural-speech recovery using the latent lens.}
Cumulative Recall@10 for the current and next words.
Parentheses report matched-frequency baselines.
Llama PI-1/3 denotes Llama-3.2 3B PI-1/3 20k (official);
Llama PI-2/3 denotes Llama-3.2 3B PI-2/3 20k; and
Qwen PI-1/3 denotes Qwen2.5 3B PI-1/3 60k.
}
\label{tab:natural_latent}
\end{table*}

\begin{table*}[t]
\centering
\small
\setlength{\tabcolsep}{4pt}
\begin{tabular}{l|ccc|ccc}
\toprule
& \multicolumn{3}{c|}{\textbf{Current word}}
& \multicolumn{3}{c}{\textbf{Next word}} \\
\textbf{Dataset}
& \textbf{Llama PI-1/3}
& \textbf{Llama PI-2/3}
& \textbf{Qwen PI-1/3}
& \textbf{Llama PI-1/3}
& \textbf{Llama PI-2/3}
& \textbf{Qwen PI-1/3} \\
\midrule
Controlled
& 87.7 (9.3) & 81.7 (13.1) & 87.1 (3.5)
& 30.6 (6.8) & 32.3 (11.4) & 31.4 (3.8) \\
\midrule
People's Speech
& 94.6 (8.0) & 94.8 (10.7) & 94.8 (2.3)
& 70.9 (6.3) & 81.5 (9.6) & 76.0 (1.7) \\
LibriSpeech
& 85.6 (8.5) & 85.4 (11.1) & 86.0 (3.1)
& 49.4 (7.4) & 52.2 (11.0) & 51.9 (2.6) \\
VoxPopuli
& 91.5 (9.1) & 91.9 (12.1) & 92.5 (3.3)
& 55.5 (7.4) & 57.4 (11.4) & 58.3 (2.3) \\
TED-LIUM
& 87.9 (8.4) & 87.9 (11.3) & 88.1 (2.7)
& 52.7 (7.2) & 56.1 (10.7) & 54.9 (1.9) \\
SWC
& 82.1 (7.3) & 81.8 (10.0) & 82.3 (3.5)
& 43.9 (6.2) & 47.2 (9.9) & 46.6 (3.0) \\
VCTK
& 97.4 (9.7) & 91.7 (11.5) & 93.2 (2.6)
& 57.6 (8.0) & 53.5 (10.9) & 54.7 (2.1) \\
Flickr8k
& 87.5 (6.8) & 81.7 (9.2) & 82.0 (3.2)
& 39.4 (4.7) & 42.8 (8.4) & 41.7 (2.5) \\
\midrule
Natural speech
& 89.4 (8.2) & 87.7 (10.6) & 88.0 (2.9)
& 57.4 (6.7) & 58.5 (9.9) & 56.7 (2.3) \\
\bottomrule
\end{tabular}
\caption{
\textbf{Natural-speech recovery using linear probes.}
Cumulative Recall@10 for the current and next words.
Parentheses report matched-frequency baselines.
}
\label{tab:natural_probe}
\end{table*}

\end{document}